\documentclass[pdflatex,sn-basic]{sn-jnl}

\usepackage{amsmath,amssymb}
\usepackage{booktabs}
\usepackage{graphicx}
\usepackage{multirow}
\usepackage{tabularx}
\usepackage{tikz}
\usepackage{xcolor}
\usetikzlibrary{arrows.meta,positioning,calc,decorations.pathreplacing}

\newcommand{\wt}[1]{#1}

\begin{document}

\title{Complex Problem Solving in Large Language Models: A Statistical Control Survey and Diagnostic Framework}

\author[1]{\fnm{Jiazhang} \sur{Cai}}
\author[2]{\fnm{Tao} \sur{Wang}}
\author[3]{\fnm{Ruidong} \sur{Zhang}}
\author[3]{\fnm{Siyuan} \sur{Li}}
\author[4]{\fnm{Terry} \sur{Ma}}
\author[2]{\fnm{Luyang} \sur{Fang}}
\author[2]{\fnm{Haoran} \sur{Lu}}
\author[5]{\fnm{Huimin} \sur{Cheng}}
\author[2]{\fnm{Yingchuan} \sur{Zhang}}
\author[2]{\fnm{Shushan} \sur{Wu}}
\author[6]{\fnm{Rui} \sur{Xie}}
\author[5]{\fnm{Lin} \sur{Tang}}
\author[7]{\fnm{Chao} \sur{Huang}}
\author[2]{\fnm{Rongjie} \sur{Liu}}
\author[2]{\fnm{Ziyu} \sur{Liu}}
\author[2]{\fnm{Meizhi} \sur{Yu}}
\author[8]{\fnm{Yongkai} \sur{Chen}}
\author[3]{\fnm{Yifan} \sur{Zhou}}
\author[7]{\fnm{Zeliang} \sur{Sun}}
\author[6]{\fnm{Chang} \sur{Liu}}
\author[3]{\fnm{Zhen} \sur{Xiang}}
\author[9]{\fnm{Wei} \sur{Xiao}}
\author[3]{\fnm{Zixin} \sur{Rao}}
\author[2]{\fnm{Xinyi} \sur{Liu}}
\author[2]{\fnm{Yutong} \sur{Hu}}
\author[10]{\fnm{Mengrui} \sur{Zhang}}
\author[11]{\fnm{Jing} \sur{Zhang}}
\author[3]{\fnm{Weidi} \sur{Luo}}
\author[12]{\fnm{Jincheng} \sur{Yu}}
\author[3]{\fnm{Zhengliang} \sur{Liu}}
\author[3]{\fnm{Weihang} \sur{You}}
\author[3]{\fnm{Hanqi} \sur{Jiang}}
\author[3]{\fnm{Yi} \sur{Pan}}
\author[3]{\fnm{Junhao} \sur{Chen}}
\author[3]{\fnm{Xinliang} \sur{Li}}
\author*[3]{\fnm{Tianming} \sur{Liu}}\email{tliu@uga.edu}
\author*[2]{\fnm{Wenxuan} \sur{Zhong}}\email{wenxuan@uga.edu}
\author*[2]{\fnm{Ping} \sur{Ma}}\email{pingma@uga.edu}

\affil[1]{\orgdiv{Department of Genetics and Genomic Science},
  \orgname{Icahn School of Medicine at Mount Sinai},
  \orgaddress{\city{New York}, \state{NY}, \country{USA}}}

\affil[2]{\orgdiv{Department of Statistics},
  \orgname{University of Georgia},
  \orgaddress{\city{Athens}, \state{GA}, \country{USA}}}

\affil[3]{\orgdiv{School of Computing},
  \orgname{University of Georgia},
  \orgaddress{\city{Athens}, \state{GA}, \country{USA}}}

\affil[4]{\orgdiv{School of Computer Science},
  \orgname{Carnegie Mellon University},
  \orgaddress{\city{Pittsburgh}, \state{PA}, \country{USA}}}

\affil[5]{\orgdiv{Department of Biostatistics},
  \orgname{Boston University School of Public Health},
  \orgaddress{\city{Boston}, \state{MA}, \country{USA}}}

\affil[6]{\orgdiv{School of Data, Mathematical, and Statistical Sciences},
  \orgname{University of Central Florida},
  \orgaddress{\city{Orlando}, \state{FL}, \country{USA}}}

\affil[7]{\orgdiv{Department of Epidemiology \& Biostatistics},
  \orgname{University of Georgia},
  \orgaddress{\city{Athens}, \state{GA}, \country{USA}}}

\affil[8]{\orgdiv{Department of Statistics},
  \orgname{Harvard University},
  \orgaddress{\city{Cambridge}, \state{MA}, \country{USA}}}

\affil[9]{\orgname{UT MD Anderson Cancer Center},
  \orgaddress{\city{Houston}, \state{TX}, \country{USA}}}

\affil[10]{\orgdiv{Quantitative Sciences Unit, Department of Medicine},
  \orgname{Stanford University School of Medicine},
  \orgaddress{\city{Stanford}, \state{CA}, \country{USA}}}

\affil[11]{\orgdiv{Department of Computer Science and Engineering},
  \orgname{University of Texas at Arlington},
  \orgaddress{\city{Arlington}, \state{TX}, \country{USA}}}

\affil[12]{\orgdiv{Statistics \& Data Science GIDP},
  \orgname{University of Arizona}, 
  \orgaddress{\city{Tucson}, \state{AZ}, \country{USA}}}

\abstract{Complex problem solving (CPS) with large language models (LLMs) is often framed as a matter of stronger reasoning or longer generation. Yet early-step error amplification, prompt brittleness, and failures to revise incorrect commitments are difficult to explain by missing knowledge or expressive capacity alone. This survey interprets CPS as a sequential estimation-and-decision problem over a latent solution state. A controller maintains a belief about an unobserved solution trajectory, updates it as noisy intermediate evidence arrives, and decides whether to commit, verify, branch, roll back, or abstain to minimize expected loss. Reasoning supplies candidate transitions and interpretations, whereas process control shapes and evaluates those proposals and regulates subsequent transitions and observations. Within this framework, we organize existing methods around five components: explicit state representation, transition structuring, validation and constraint enforcement, search and rollback, and uncertainty management. We also interpret evaluation metrics according to the statistical quantities they estimate. The framework further yields a diagnostic hypothesis: interventions should be most effective when they target the error or uncertainty component implicated by an observed failure. We distinguish systematic, stochastic, and irreducible error together with epistemic and aleatoric uncertainty, and call this alignment \emph{problem--control fit} and its failure \emph{control mismatch}. For example, additional sampling may reduce sampling variability while leaving a shared systematic error unchanged. This perspective clarifies what current methods estimate and control, what remains uncontrolled, and why reliable validation, targeted recovery, calibrated uncertainty, and matched-budget evaluation are central open problems.}


\maketitle

\section{Introduction}

Complex problem solving (CPS) with large language models (LLMs) is commonly framed as a question of stronger reasoning or longer chains of generation.
However, many persistent failure modes, such as early-step error amplification, brittleness to prompt variation, and the inability to revise incorrect commitments, are difficult to explain purely in terms of missing knowledge or insufficient expressive capacity.
In this survey, we argue that these failures are better understood as failures of process control.
From this perspective, CPS requires maintaining correctness over extended trajectories under uncertainty, rather than producing locally plausible intermediate steps.
Effective CPS systems must therefore maintain task-relevant intermediate state, externalize control-critical parts when inspection or intervention is needed, constrain and structure state transitions, support exploration and rollback, validate intermediate claims, and manage uncertainty throughout execution.
We use this control-centric lens to reinterpret a wide range of LLM-based methods, analyze which aspects of the problem-solving process they actually control, and identify what remains systematically uncontrolled.

As LLMs are increasingly deployed in settings that demand multi-step deliberation, such as planning, analysis, and tool-mediated decision making, their limitations are no longer dominated by isolated factual errors.
Instead, failures often arise from an inability to keep a long-horizon process on track.
Small local mistakes, ambiguous intermediate assumptions, or unverified transitions can accumulate into globally invalid outcomes, even when each individual step appears reasonable in isolation.
This suggests that CPS should not be viewed primarily as a matter of generating more detailed intermediate reasoning, but as a matter of controlling how a solution trajectory evolves.
Under this view, the central challenge is not whether the model can propose plausible next steps, but whether the system can constrain, monitor, and correct those steps as uncertainty unfolds over time.

Despite notable progress on benchmark reasoning tasks, current LLM-based systems remain fragile on CPS problems that require long-horizon consistency.
A recurring pattern is that once an early intermediate decision is incorrect, subsequent steps tend to remain locally coherent while reinforcing the initial error~\citep{creswell2022faithfulreasoningusinglarge, lightman2023let}.
This behavior reflects a lack of effective rollback and revision mechanisms, rather than a lack of syntactic or semantic fluency.
In addition, many systems commit prematurely to a single trajectory and lack structured means to explore alternatives or compare competing hypotheses.
Finally, intermediate claims are often weakly validated or not validated at all, allowing incorrect assumptions to persist until they irreversibly contaminate downstream reasoning.
Taken together, these observations suggest that many CPS failures can be read as missing or insufficient control rather than as deficiencies of reasoning capability per se. A central implication is that interventions should be chosen in light of the control bottleneck implicated by the observed failure rather than defaulting to more reasoning. Organizing methods by the control component they implement therefore also suggests a diagnostic procedure: ask which component is plausibly failing before asking which reasoning method raises accuracy. We develop this diagnostic view, and the failure mode we call \emph{control mismatch}, in Section~\ref{sec:framework}.

In this survey, we use CPS to refer to problem-solving settings that require maintaining task-relevant intermediate state, respecting nontrivial constraints on state transitions, and remaining robust to uncertainty and partial observability. Whether that state is fully explicit, partly externalized, or partly latent is a system-design choice rather than part of the task definition.
We draw a conceptual distinction between reasoning and control that we state statistically throughout the survey. Reasoning denotes the base proposal mechanism that, given a current state and partial observations, generates candidate transitions or interpretations from learned associations. Process control comprises the mechanisms that shape or gate those proposals, update a belief over the latent solution state as evidence arrives, and choose whether to commit, verify, branch, roll back, or abstain.
This distinction separates fluent generation from reliable execution: long chains of text alone do not constitute CPS unless the trajectory is managed by mechanisms that constrain, evaluate, and revise it over time. Accordingly, we focus on methods that introduce process-level structure and oversight, rather than treating extended chain-of-thought generation as a sufficient indicator of CPS ability.

LLMs create new opportunities for CPS not because they inherently solve long-horizon control problems, but because they can be embedded within systems that externalize and enforce control.
Intermediate states can be made explicit through structured representations, transitions can be constrained via planning or program-like decompositions, and validation can be delegated to tools with executable or retrieval-based semantics.
Agentic architectures further enable the separation of proposal, execution, and verification roles, introducing feedback loops that are absent in single-pass generation.
At the same time, these opportunities come with significant limitations.
Control signals may be noisy or misaligned, validators can be exploited or overfitted, and recovery mechanisms often degrade into repeated retries without principled revision.
As a result, the mere presence of scaffolding does not guarantee effective control, and understanding its failure modes is essential for assessing CPS systems.

This survey advances a statistical interpretation of CPS in LLM-based systems.
We argue that many methods commonly described as improvements in reasoning are better understood as implementing one component of a recursive estimation-and-decision loop, and that reading them through classical statistical objects, a state-space model, a Bayes filter, a bias--variance decomposition, and a sequential decision rule, clarifies what each method estimates and controls and where gaps remain.
Beyond classifying methods, we use these objects diagnostically: we introduce the notion of \emph{problem--control fit}, decompose the error of a solved trajectory into systematic, stochastic, and irreducible components, and summarize the reading in a diagnostic table (Table~\ref{tab:tab1}) that pairs characteristic failure signatures with plausibly implicated error and control components, candidate matched interventions, and interventions likely to leave the observed failure unaddressed.
We further analyze evaluation practices through the lens of control coverage and process faithfulness, highlighting why outcome-only metrics often fail to diagnose CPS failures.
Finally, we identify recurring sources of uncontrolled behavior, including weak rollback policies, brittle validation, and poor uncertainty calibration, as central obstacles to reliable CPS.

Our organizing principle also distinguishes this survey from prior overviews, which typically group methods by capability or technique pattern: surveys of chain-of-thought and prompting~\citep{sahoo2024systematic}, in-context learning~\citep{dongSurveyIncontextLearning2024}, tool-augmented language models~\citep{mialon2023augmentedlanguagemodelssurvey}, self-correction~\citep{pan2023automatically}, uncertainty quantification~\citep{shorinwa2024survey, huang2024survey}, efficient and test-time reasoning~\citep{qu2025surveyefficientreasoninglarge, snell2024scaling}, and evaluation~\citep{chang2024survey}. We instead organize the literature around the statistical role each method plays in a single estimation-and-decision loop, and we go one step further by using that structure diagnostically. Many techniques described as reasoning improvements are better understood as estimating, shaping, or controlling one component of this loop, and which technique helps should depend on the failure component actually implicated in a given setting. This lets us treat methods, evaluation, and open challenges under a single set of axes, and lets us ask not only which method raises accuracy but which statistical failure it is positioned to repair.

We make three contributions. First, we formulate CPS as sequential estimation and decision over a latent solution state, using an action-conditioned Bayes-filter scaffold with a decision layer as the spine of the survey. Second, we reinterpret major LLM CPS methods as components of that object, including proposal-shaping mechanisms, likelihood surrogates, variance-reduction estimators, posterior-search procedures, and decision rules, rather than as an unordered catalogue of techniques. Third, we turn the framework into a diagnostic hypothesis by decomposing trajectory error into bias, variance, and noise and predictive uncertainty into epistemic and aleatoric parts, and by linking characteristic failure signatures to plausible control bottlenecks and matched interventions.

\section{From Reasoning to Process Control}
We first review the reasoning-centric view that dominates the literature and show where it breaks down, then develop the process-control framework that organizes the rest of the survey.

\subsection{The Reasoning-Centric View and Its Limits}

Much of the literature on LLM problem solving is framed around reasoning as the primary explanatory construct.
Under this view, solving complex tasks amounts to producing a sequence of intermediate steps that are locally coherent and logically connected, with errors attributed to insufficient reasoning depth or missing knowledge.
There are several implicit assumptions under this framing: intermediate steps need not be explicitly represented as state, transitions between steps can be evaluated implicitly by the model itself, and correctness at the token or step level is a reasonable proxy for correctness of the overall solution.
While these assumptions may be adequate for short-horizon tasks, they become increasingly fragile as problem-solving trajectories grow longer and more uncertain.

Chain-of-Thought (CoT) prompting is often presented as a mechanism for exposing intermediate reasoning and improving multi-step performance.
However, the presence of an explicit textual trace does not guarantee that the underlying problem-solving process is controlled.
CoT primarily externalizes intermediate tokens, but it does not, by itself, enforce constraints on how those tokens relate to a well-defined state or whether transitions between steps are valid.
As a result, chain-of-thought can produce explanations that are fluent and internally consistent while remaining causally disconnected from the final answer.
In CPS settings, such unverified intermediate steps can introduce silent failure modes, where early mistakes propagate through the trajectory without triggering corrective mechanisms.

Another influential line of work attributes improvements in complex reasoning to emergent abilities arising from scale.
Empirically, larger models often exhibit better performance on multi-step benchmarks, which is sometimes interpreted as evidence that sufficient scale can implicitly internalize planning, verification, and error correction.
However, emergence through scale does not imply explicit control.
Even large models can remain overconfident and may fail to revise incorrect reasoning without reliable external feedback; these findings concern specific models, prompts, and tasks rather than a universal inability to self-correct~\citep{huang2024selfcorrect,stechly2024self,kadavath2022language}.
These behaviors suggest that scale can improve the proposal of candidate transitions, but does not reliably supply mechanisms for monitoring, validating, and revising long-horizon processes.

A central limitation of reasoning-centric views is their emphasis on token-level or step-level correctness.
In CPS, the primary objective is not that each intermediate step appears reasonable in isolation, but that the overall trajectory remains stable under uncertainty.
This requires maintaining task-relevant intermediate state, externalizing control-critical parts when needed for inspection or intervention, constraining allowable transitions, and incorporating feedback signals that can trigger revision or rollback.
Without such mechanisms, fluent reasoning traces can mask fundamental instabilities, leading to brittle behavior when tasks deviate from benchmark conditions.
These observations motivate a shift from reasoning as generation toward problem solving as controlled execution, which forms the basis for the control-centric framework developed in the remainder of this section.

\subsection{Error Propagation as Evidence of Missing Control}
The reasoning-centric view struggles to explain a failure mode that is central to CPS: small errors that compound across steps. Read as recursive estimation, this compounding is the behavior of a state estimate under composition. The reasoning state is not in general a differentiable vector, so the following is an analytical scaffold rather than an empirical claim: conceptually, and locally around a reference trajectory on which the state admits a numerical embedding, let $\mathbf{e}_t = \hat{\mathbf{s}}_t - \mathbf{s}_t^\star$ denote the deviation of the estimated state from the correct one at step $t$, which to first order evolves as
\begin{equation}
\mathbf{e}_{t+1} = \mathbf{J}_t\, \mathbf{e}_t + \boldsymbol{\varepsilon}_t,
\qquad \mathbf{J}_t = \left.\frac{\partial \boldsymbol{f}}{\partial \mathbf{s}}\right|_{\mathbf{s}_t},
\label{eq:error-recursion}
\end{equation}

where $\boldsymbol{f}$ is the effective per-step update under the selected control action and $\boldsymbol{\varepsilon}_t$ is the noise injected at step $t$. When $\|\mathbf{J}_t\|\leq\kappa<1$, the local update is contractive and attenuates perturbations. By contrast, $\|\mathbf{J}_t\|>1$ means that the update permits expansion in at least one direction; whether a particular error grows depends on its alignment with those directions and on the product of Jacobians along the trajectory. Under this local scaffold, validation can reduce newly injected error and rollback can limit the effect of an unstable sequence of updates, but neither mechanism guarantees global contraction. This motivates the framework developed below.

\begin{figure}[t]
    \centering
    \includegraphics[width=0.72\linewidth]
        {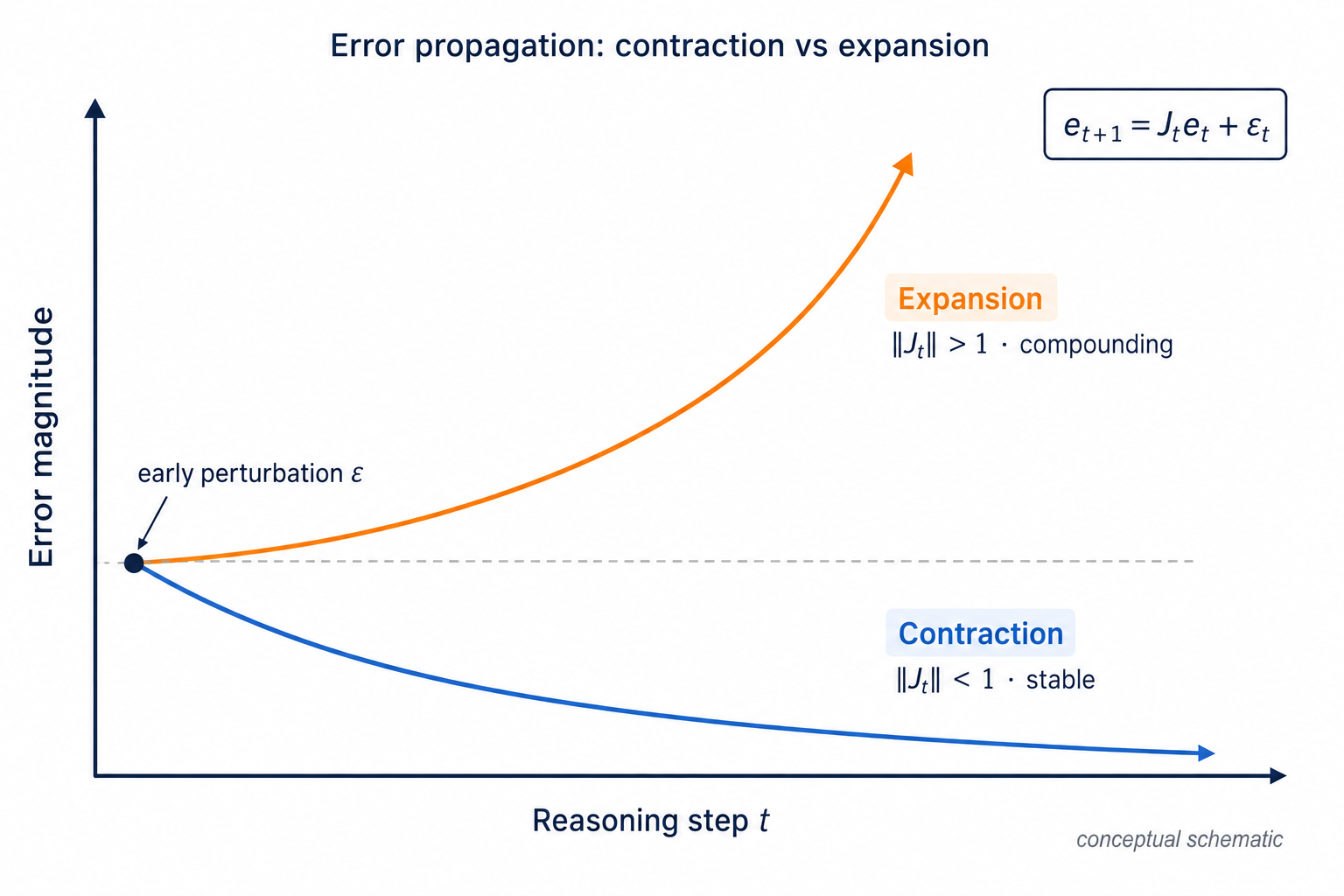}
    \caption{Error propagation under contracting and expanding updates. Starting from the same early perturbation, a locally contracting update ($\lVert J_t\rVert<1$) attenuates the error, whereas an expanding update ($\lVert J_t\rVert>1$) amplifies it geometrically along the trajectory. The figure is an analytical schematic of the local stability interpretation in Equation~\eqref{eq:error-recursion}, rather than an empirical trajectory}
    \label{fig:fig1}
\end{figure}

As illustrated in Figure~\ref{fig:fig1}, because later steps condition on earlier outputs, an early numerical, logical, or interpretive error can influence the remainder of a multi-step trajectory; this pattern is documented in mathematical and compositional reasoning settings~\citep{creswell2022faithfulreasoningusinglarge,lightman2023let}. Within the analytical scaffold of Equation~\eqref{eq:error-recursion}, numerical or symbolic slips contribute to the injected-error term $\boldsymbol{\varepsilon}_t$; reasoning from a false premise can place the trajectory in a locally unstable regime; and instruction misinterpretation can be represented as a biased update or incorrect initial state. These are interpretive correspondences, not claims that the cited systems expose a differentiable state or Jacobian~\citep{wei2022chain,yao2023tree}.

Existing mitigations intervene at different points in this scaffold: self-correction and reflection revisit earlier outputs~\citep{madaan2023self,shinn2023reflexion}; multi-agent interaction can expose disagreements, although correlated agents may share the same error~\citep{li2023camel,hong2024metagpt,wu2024autogen}; tool integration can make selected operations executable~\citep{gao2023pal}; and process supervision supplies step-level training or scoring signals~\citep{lightman2023let}. These mechanisms can reduce particular failure modes, but the cited results do not establish that the overall reasoning dynamics are contractive.

\subsection{Complex Problem Solving as Controlled Trajectory Execution}

We formalize complex problem solving as the execution of a trajectory through a space of intermediate states under uncertainty.
A problem instance defines an initial state, a set of constraints over admissible states and transitions, and a termination condition that specifies when a solution is acceptable.
Solving the problem requires selecting and executing a sequence of state transitions such that the resulting trajectory satisfies global constraints, even when intermediate observations are partial, noisy, or ambiguous.
Under this formulation, failure arises not only from choosing an incorrect transition, but from the inability to detect, correct, or recover from such choices once they occur.

Concretely, we treat CPS as a partially observed controlled state-space model. A latent state $s_t \in \mathcal{S}$ carries the problem-relevant information at step $t$. Given a control action $a_t$, an algorithm-induced transition kernel $q(s_{t+1}\mid s_t,a_t)$ describes how candidate proposals together with the selected control operation move the represented solution state, and an observation model $p(o_{t+1}\mid s_{t+1},a_t)$ describes the evidence, such as a validator score, execution result, or tool response, made available by that operation. The kernel $q$ is therefore a model of the solver's controlled state evolution, not a claim about physical environment dynamics. Because $s_t$ is not directly observed, the controller maintains an idealized belief $b_t(s)=p(s_t=s\mid o_{1:t},a_{0:t-1})$, updated recursively through
\begin{equation}
\bar b_{t+1}(s) = \int q(s\mid s',a_t)\, b_t(s')\, ds',
\qquad
b_{t+1}(s) \propto p(o_{t+1}\mid s,a_t)\, \bar b_{t+1}(s).
\label{eq:bayes-filter}
\end{equation}
The first step propagates the belief under the selected action, and the second reweights it by the evidence that action exposes. We distinguish this latent problem state $s_t$ from any externalized reasoning trace or structured interface $r_t$ and from the controller belief $b_t$: a chain-of-thought trace can make parts of the process inspectable without being a faithful readout of either $s_t$ or $b_t$. This is an idealized recursion: current models do not represent $b_t$ explicitly or update it by Bayes' rule, so we use the filter as an analytical backbone onto which the remainder of the survey maps individual methods, not as a description of how any system literally computes. The statistical lens does not replace the control framing; it specifies what the control loop estimates and which decision problem it solves.

This perspective departs from reasoning-centric views that emphasize the correctness of individual steps or inferences.
In CPS settings, local plausibility does not guarantee global validity: intermediate decisions may appear reasonable in isolation while steering the trajectory toward states from which recovery is difficult or impossible.
Consequently, CPS fundamentally concerns maintaining trajectory-level stability, rather than producing a sequence of locally coherent steps.

Within this framework, large language models are best understood as components that propose candidate transitions or interpretations based on learned patterns.
Given a current state and partial observations, an LLM can generate plausible next actions, decompositions, or hypotheses.
However, proposal alone does not constitute control.
The system must still determine whether a proposed transition is admissible, whether it violates known constraints, and whether uncertainty warrants exploration of alternatives rather than commitment.

Standard autoregressive LLM generation does not, by default, expose a persistent task state or guarantee enforcement of global constraints over an extended trajectory. Information may remain available in the token context or hidden activations, but neither constitutes a reliably typed, updateable state or a calibrated signal for deciding when an earlier commitment should be revised. Consequently, CPS systems often add explicit state, validation, search, or other structured control mechanisms around the base model.

Under a control-centric view, solving a CPS problem is not equivalent to generating a convincing explanation or arriving at a correct final answer by chance.
A solution is characterized by the existence of a controlled execution process that can reliably reach acceptable terminal states across variations in inputs, observations, and intermediate uncertainties.
This shifts the evaluation focus from outcome-only correctness to the properties of the underlying process, including whether control-relevant state is represented in a form suitable for the needed inspection or intervention, whether transitions are constrained and verifiable, and whether the system can recover from errors without degenerating into repeated retries or arbitrary restarts.

This formulation highlights why many existing LLM-based approaches succeed on narrow benchmarks yet fail under distribution shift or increased task complexity.
Without explicit control mechanisms, performance gains often reflect improved proposal quality rather than improved trajectory management.
These observations motivate analyzing CPS methods in terms of the control components they introduce, which we develop in the next subsection.

\subsection{The Estimation-and-Decision Loop and Its Diagnostic}
\label{sec:framework}

The preceding discussion motivates a vocabulary that we use throughout the rest of the survey. We fix three terms. \emph{Reasoning} denotes the base proposal step, the component that, given a current state and partial observations, proposes candidate transitions or interpretations from learned associations. \emph{Process control} denotes the mechanisms that shape or gate those proposals, recursively update beliefs from evidence, and select actions that determine whether the system commits, verifies, branches, rolls back, or abstains. \emph{Complex problem solving} denotes the class of problem settings that require such state maintenance and feedback-sensitive decisions to be solved reliably. For exposition, we use a one-step Bayes decision layer: at each step the controller selects an action $a_t$ from $\{\textsf{commit},\textsf{verify},\textsf{branch},\textsf{rollback},\textsf{abstain}\}$ to minimize the expected loss under the current belief,
\begin{equation}
a_t^\star = \arg\min_{a}\ \mathbb{E}_{s\sim b_t}\bigl[L(s,a)\bigr].
\label{eq:bayes-decision}
\end{equation}
In these terms, reasoning proposes; control shapes those proposals, updates the belief, and commits, checks, explores, or withdraws. The action-conditioned recursion in \eqref{eq:bayes-filter} makes this feedback explicit: control affects not only the final choice but also which transitions are taken and which observations become available.

\begin{figure}[t]
    \centering
    \includegraphics[width=0.92\linewidth]
        {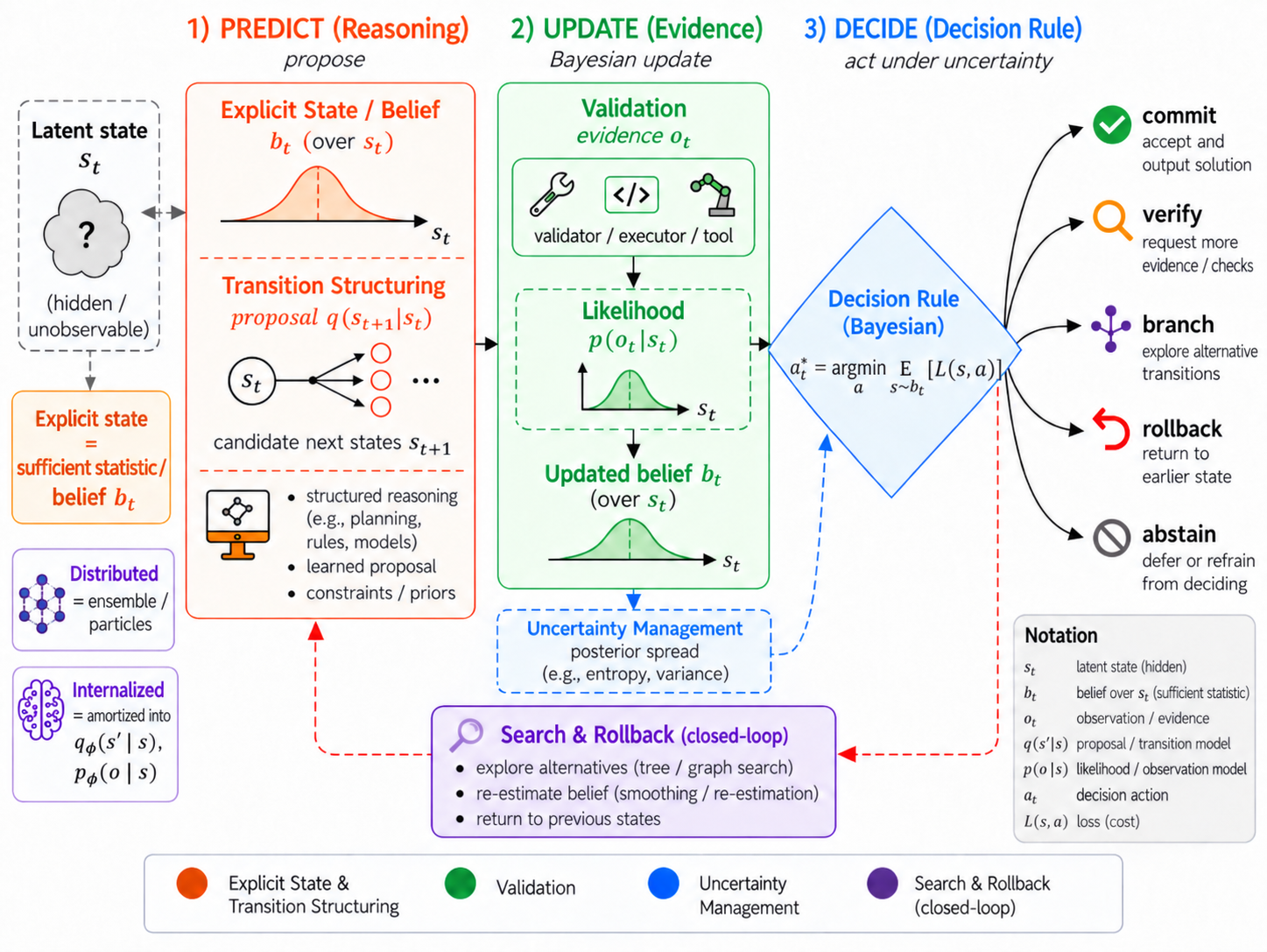}

    \caption{The estimation-and-decision loop. Reasoning proposes candidate moves, while the selected control action shapes the next-state transition and the evidence that becomes available; the observation model then updates the belief, and the decision rule commits, verifies, branches, rolls back, or abstains under that belief. Decision-relevant uncertainty, not posterior spread alone, informs how aggressively the rule explores or commits. The five components of Section~\ref{sec:framework} are parts of this one loop, and distributing control across agents and internalizing it at training time are cross-cutting dimensions that reshape, but do not replace, it}
    \label{fig:fig2}
\end{figure}

We organize the loop into five components and two cross-cutting dimensions, each identified with a part of the filter in \eqref{eq:bayes-filter} and the decision rule in \eqref{eq:bayes-decision}. The components are (i) \emph{explicit state representation}, the externalized statistic or interface through which control-relevant state can be inspected and manipulated, ideally a sufficient one for the downstream decision; (ii) \emph{transition structuring}, how the base proposal is constrained or reshaped into the controlled transition kernel $q(s_{t+1}\mid s_t,a_t)$; (iii) \emph{validation and constraint enforcement}, the action-conditioned observation model $p(o_{t+1}\mid s_{t+1},a_t)$ that reweights the belief; (iv) \emph{search and rollback}, the handling of competing posterior modes through exploration and re-estimation; and (v) \emph{uncertainty management}, the decision-relevant characteristics of the belief, including dispersion and multimodality, that inform the action rule together with the loss. As illustrated in Figure~\ref{fig:fig2}, these form a closed loop rather than an independent checklist: the proposal and selected action advance the belief, validation reweights it, low support for the current trajectory can motivate search or rollback, and decision-relevant uncertainty informs how aggressively the rule explores or commits. This closed loop is a design ideal rather than a description of current practice: the representative methods reviewed below usually emphasize a subset of the components and leave others implicit or externally supplied. That the loop is rarely closed is itself one of our central observations. Two further dimensions cut across this loop: control may be \emph{distributed} across multiple agents, and control may be \emph{internalized} into model weights at training time rather than supplied by external scaffolding.

This framework provides the spine for the remainder of the survey. Section~\ref{sec:taxonomy} organizes methods by the component they implement, Table~\ref{tab:tab2} maps representative methods onto these components, Section~\ref{sec:benchmarks} maps evaluation practice onto the same axes, and Section~\ref{sec:challenges} revisits each component to ask where control remains missing.

\begin{figure}[t]
    \centering
    \includegraphics[width=0.92\linewidth]
        {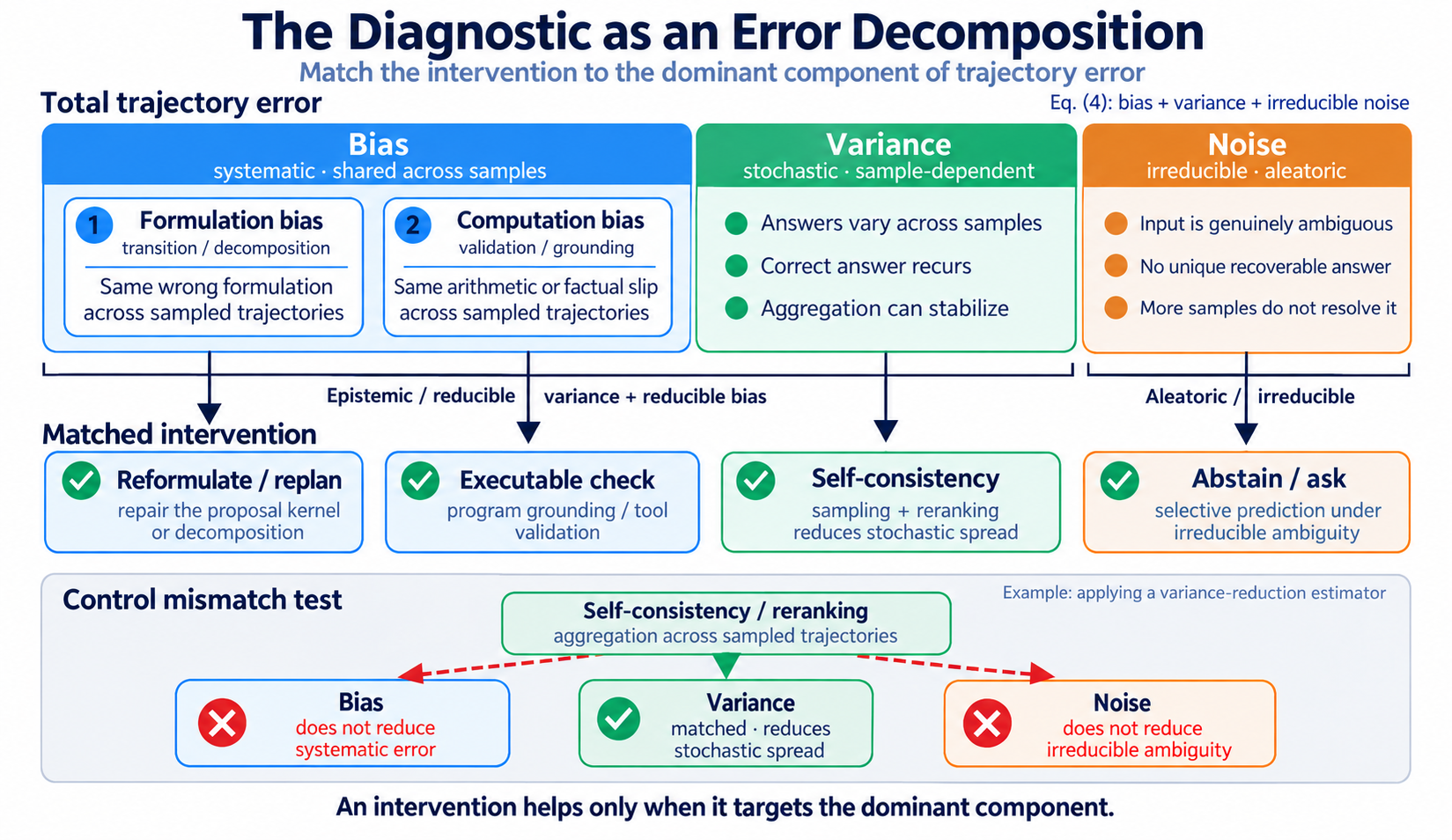}
    \caption{The diagnostic as an error decomposition. Following~\eqref{eq:bias-var-noise}, the error of a solved trajectory separates into a systematic component (bias, shared across sampled trajectories), a stochastic component (variance, sample-dependent), and an irreducible component (noise, aleatoric). The systematic component splits into a formulation bias, repaired at the transition by reformulation or replanning, and a computation bias, repaired at validation by an executable check. \wt{Sampling-based aggregation primarily targets sample-dependent variation; it may leave errors shared by the dominant trajectories and irreducible ambiguity unchanged, which is the control mismatch illustrated at the bottom}}
    \label{fig:fig3}
\end{figure}

Reading the loop statistically has an immediate diagnostic consequence. Decompose the expected error of a solved trajectory into three parts,
\begin{equation}
\mathbb{E}\bigl\|\hat y - y^\star\bigr\|^2
= \underbrace{\bigl\|\mathbb{E}[\hat y] - y^\star\bigr\|^2}_{\text{bias (systematic)}}
+ \underbrace{\mathrm{Var}(\hat y)}_{\text{variance (stochastic)}}
+ \underbrace{\sigma^2}_{\text{noise (irreducible)}},
\label{eq:bias-var-noise}
\end{equation}
where $\hat y$ is the answer the process returns. The decomposition is exact for squared-error loss; for the discrete answers typical of CPS we use it in the generalized sense of \citet{domingos2000unified}, where analogous systematic, variance, and noise terms exist but need not add linearly, and we rely only on the qualitative separation between a systematic term that resampling cannot reduce and a variance term that it can. Different interventions target different observable failure modes, so the useful question is not only which method raises accuracy, but which component of error it is positioned to address. We call it a \emph{problem--control fit} when an intervention targets the error source plausibly implicated by the observed failure, and a \emph{control mismatch} when it targets a different component. A failure signature does not uniquely identify its cause; the mapping below is therefore a diagnostic hypothesis rather than an identification result. The interventions are not interchangeable: aggregation across sampled trajectories primarily targets sampling variability, but it need not correct an error shared by the dominant trajectories or resolve irreducible ambiguity. Control mismatch can therefore arise when additional sampling is applied to a persistent formulation error, or when longer reasoning is allocated to a step that was not the bottleneck; in such cases, cost can increase without addressing the observed failure. This is not only a conceptual worry: recent studies of over-thinking find that lengthening reasoning or spending more inference-time compute can leave accuracy flat or reduce it~\citep{chen2025think23overthinkingo1like, wang2025thoughtsplaceunderthinkingo1like}. These results do not by themselves isolate the mismatch mechanism, since they do not hold the added computation fixed while varying the component it targets, so we treat the strong reading, that misdirected effort actively hurts, as a prediction of the framework rather than an established fact. Table~\ref{tab:tab1} states this view directly, pairing characteristic failure signatures with plausibly implicated error components, candidate matched interventions, and interventions that are likely to add cost without addressing that signature; Figure~\ref{fig:fig3} illustrates the decomposition together with the mismatch hypothesis. The remainder of the survey returns to this diagnostic reading: each method subsection in Section~\ref{sec:taxonomy} begins by naming the problems whose bottleneck lands on its component and ends by noting what that component cannot control.

\begin{table}[t]
\centering
\footnotesize
\setlength{\tabcolsep}{4pt}
\caption{Problem--control fit as a diagnostic hypothesis, read through the error decomposition of \eqref{eq:bias-var-noise}. Each row pairs a characteristic failure signature with a plausibly implicated error component (and loop component), a candidate matched intervention, and an intervention likely to leave that signature unaddressed. The mapping is not unique or causally identified from the signature alone. \wt{Sampling-based aggregation primarily addresses sampling variability and may leave shared systematic errors or irreducible ambiguity unchanged.}}
\label{tab:tab1}
\begin{tabularx}{\textwidth}{@{}p{3.0cm}p{2.9cm}XX@{}}
\toprule
Failure signature & Likely implicated term (component) & Candidate matched intervention & Likely mismatch \\
\midrule
Same wrong formulation across samples & Formulation bias (transition) & Reformulate, replan & Self-consistency \\
Same arithmetic or factual slip across samples & Computation bias (validation) & Executable check, program grounding & Longer CoT \\
Answer varies, correct one recurs & Variance (search) & Self-consistency, reranking & Longer single chain \\
Confident on ambiguous input & \wt{Input ambiguity / aleatoric uncertainty} & \wt{Clarify, abstain, or use selective prediction} & Force an answer \\
Early error contaminates the rest & State-estimate divergence (validation, state) & Step validation, targeted rollback & Resample whole trajectory \\
Early commitment forecloses paths & Posterior collapsed early (search) & Branch, keep modes & Validate one trajectory \\
Facts or dependencies lost & Insufficient statistic (state) & Typed belief state & More agents, longer traces \\
Overconfident, miscalibrated & \wt{Calibration / decision error (uncertainty)} & Recalibrate, conformal, proper scoring & Sharpen further \\
\bottomrule
\end{tabularx}
\end{table}

These signatures correspond to familiar problem types, and the diagnostic view associates each with a plausible error component in the sense of \eqref{eq:bias-var-noise}, with the caveat that the same signature may have multiple causes and a single problem usually stresses several components at once, so the mapping should be read as a hypothesis about an observed failure rather than as a fixed label on a domain. In arithmetic and multi-step calculation, the reasoning is often short but brittle to a single slip, so validation dominates: grounding the computation in executable code, as in program-aided prompting, improves accuracy over longer natural-language derivations~\citep{gao2023pal, chen2023pot}. Code generation is validation-dominated as well, but through execution against tests, where functional correctness is decided by running the program rather than by judging its explanation~\citep{chen2021evaluating}. In theorem proving and symbolic logic, validation is definitional, since a candidate proof counts only when a formal checker accepts it, and current systems place the language model inside a verifier-in-the-loop pipeline~\citep{polu2020generative, xin2024deepseek}. Math word problems and multi-hop question answering shift the burden toward transition structuring and state, because the typical failure is a wrong decomposition or a lost intermediate quantity; explicit decomposition reduces such missing-step errors~\citep{zhou2023leasttomostpromptingenablescomplex, wang2023planandsolvepromptingimprovingzeroshot}. Additional stochastic samples from the same model and prompt need not correct a decomposition error that recurs across the dominant trajectories. Planning and agentic tasks are dominated by search and rollback, since early commitments constrain later feasibility and the system must be able to revise a plan when execution contradicts it~\citep{yao2023tree, hao2023reasoning, erdogan2025planandactimprovingplanningagents}. Open-ended, commonsense, and underspecified questions are dominated by uncertainty, because there may be no single correct path and current models tend to remain overconfident and poorly calibrated in exactly these settings~\citep{liu2024uncertainty, zhang2025token}. The same lens explains why a fix that helps one failure mode may not help another: self-consistency is most useful when sampling exposes a recurring correct answer, but it provides no guarantee against a shared systematic error~\citep{wang2022self}; a symbolic checker can settle whether a proof satisfies its formal specification~\citep{polu2020generative}, but not whether an underspecified question has been interpreted as intended. These correspondences are tendencies, not partitions: code generation also benefits from search over candidate programs, and math word problems from validating the final computation, so a problem typically stresses two or three components at once, and the diagnostic applies to the specific failure observed rather than to the domain as a whole.

A single example makes the diagnostic concrete. Consider a model that sets up the right equations for a multi-step word problem but makes the same arithmetic slip in several sampled attempts. Repetition across samples suggests a shared computational failure rather than sampling variation alone, so the diagnostic points to validation rather than broader search. Executing a correctly formulated program can prevent the arithmetic operation itself from being performed in free-form text~\citep{gao2023pal}; it does not, however, guarantee that the model translated the problem into the right program. Additional sampling offers no comparable guarantee when the same slip dominates the sampled trajectories~\citep{wang2022self}, and a longer trace adds deliberation without directly checking the computation. The framework turns this observation into a routine diagnostic: read the failure signature, locate the implicated component, and match the intervention to it.

The same pattern appears in published results rather than only in constructed examples, and Table~\ref{tab:tab3} collects three cases read through the decomposition. Self-consistency raises PaLM-540B accuracy from $56.5$ to $74.4$ on GSM8K and from $79.0$ to $80.7$ on CommonsenseQA~\citep{wang2022self}. The task-dependent difference is consistent with aggregation being more useful in some sampling distributions than others, but the benchmark comparison alone does not identify a bias--variance decomposition. On GSM-HARD, which stresses arithmetic with larger numbers, Codex obtains $23.1$ with chain-of-thought and $61.2$ with program-aided prompting~\citep{gao2023pal}. This supports executable computation as a useful intervention for large-number arithmetic, but it does not establish the correlation structure of errors across sampled traces. Separate over-thinking studies report settings in which lengthening reasoning or spending more inference-time compute leaves accuracy flat or lower~\citep{chen2025think23overthinkingo1like,wang2025thoughtsplaceunderthinkingo1like}. Together, these findings motivate the diagnostic, but they are not a controlled matched-budget test of its causal interpretation.

\begin{table}[t]
\centering
\scriptsize
\setlength{\tabcolsep}{2pt}
\renewcommand{\arraystretch}{1.15}
\caption{Representative LLM methods mapped onto the control components they implement. Most methods realize more than one component, which is why we organize the survey by component rather than by method.}
\label{tab:tab2}

\begin{tabularx}{\textwidth}{@{}>{\raggedright\arraybackslash}Xccccccc@{}}
\toprule
Method & State & Trans. & Search/ & Valid. & Uncert. & Multi- & Train. \\
       &       &        & rollback &        &         & agent  &        \\
\midrule
Chain-of-Thought~\citep{wei2022chain}
& \checkmark & & & & & & \\

Scratchpad~\citep{nye2021show}
& \checkmark & & & & & & \\

Tree-of-Thoughts~\citep{yao2023tree}
& \checkmark & & \checkmark & & & & \\

Least-to-Most~\citep{zhou2023leasttomostpromptingenablescomplex}
& & \checkmark & & & & & \\

PAL / PoT~\citep{gao2023pal,chen2023pot}
& & \checkmark & & \checkmark & & & \\

ReAct~\citep{yao2023react}
& & \checkmark & & \checkmark & & & \\

Self-Consistency~\citep{wang2022self}
& & & \checkmark & & \checkmark & & \\

Self-Refine / Reflexion~\citep{madaan2023self,shinn2023reflexion}
& & & \checkmark & \checkmark & & & \\

Verifier-guided search~\citep{yu2024ovm}
& & & \checkmark & \checkmark & \checkmark & & \\

Process Reward Models~\citep{lightman2023let}
& & & & \checkmark & & & \checkmark \\

Multi-agent~\citep{hong2024metagpt,wu2024autogen}
& & \checkmark & \checkmark & \checkmark & & \checkmark & \\

RLHF / process supervision~\citep{ouyang2022training,setlur2024rewarding}
& & & & \checkmark & & & \checkmark \\
\bottomrule
\end{tabularx}

\par\smallskip
\parbox{\linewidth}{\scriptsize
Trans. = Transition; Valid. = Validation;
Uncert. = Uncertainty; Train. = Training.
}
\end{table}

\subsection{Scope and Limitations of This Survey}
We focus on process-level structure and oversight in language-model-based CPS, and we organize the literature around the five control components and two cross-cutting dimensions defined above. Several boundaries follow from this choice. We treat reasoning quality only insofar as it interacts with control, and we do not attempt a comprehensive account of pretraining, architectures, or model internals except where they internalize control. Multimodal and embodied settings enter through their control requirements rather than as a domain survey in their own right.

Three caveats apply to the framework itself. First, the five-component decomposition is a modeling choice rather than a unique partition; many methods realize several components at once, as Table~\ref{tab:tab2} makes explicit, so assigning a method to a component reflects where its main control contribution lies rather than a strict boundary. Second, the diagnostic in Table~\ref{tab:tab1} is a hypothesis synthesized from the literature rather than a controlled result: it predicts which component a failure signature implicates and which intervention should help, and although individual predictions are supported by the works we cite, the mapping as a whole invites systematic empirical testing, which we revisit as an open problem in Section~\ref{sec:challenges}. Third, the probabilistic objects we use, the belief $b_t$, the controlled transition kernel $q$, and decision-relevant uncertainty summaries form a decision-theoretic scaffold rather than a claim that current models compute calibrated posteriors. The scores an LLM emits are typically miscalibrated, and part of the statistical reading is to ask which methods repair the gap between a heuristic score and a decision-valid quantity. We therefore treat calibration, proper scoring, and conformal procedures as a lens for reviewing existing methods rather than as methods we introduce, consistent with the survey's aim of re-viewing the field rather than proposing new machinery.

Two further points bound the claims we make. The framework is falsifiable in one concrete respect: it predicts that, at a fixed computational budget, effort aimed at the component a failure implicates should outperform equal effort aimed elsewhere. A controlled comparison that found no such advantage, or that found added reasoning helping uniformly regardless of the failure type, would count as evidence against the diagnostic; a systematic matched-budget test is therefore a central experiment invited by the framework. The lens also has a boundary: it addresses failures of process, not failures of content. When a problem fails because the required knowledge is simply absent, because the task is genuinely ambiguous with no admissible resolution, or because it is creative or ill-posed in a way that no verifier or state can capture, the control vocabulary names the symptom without locating a fixable bottleneck.

A stronger counter-view deserves explicit mention. The recent wave of reasoning models trained with reinforcement learning on verifiable rewards, such as DeepSeek-R1, shows that some validation and revision behavior can be internalized into the policy itself rather than supplied by external scaffolding~\citep{guo2025deepseek}. If such training reliably internalized the control components, external control would become largely optional and our framing would reduce to a description of pre-RL systems. We take this possibility seriously, while noting that visible reasoning traces do not by themselves reveal the model's internal mechanism and that longer internalized reasoning is not uniformly beneficial, as the over- and under-thinking results above illustrate. The open question, which we return to in Section~\ref{sec:challenges}, is whether internalized control remains verifiable and reliable under distribution shift.

\section{Method Taxonomy via Loop Components} \label{sec:taxonomy}

\begin{figure}
    \centering
    \includegraphics[width=0.8\linewidth]{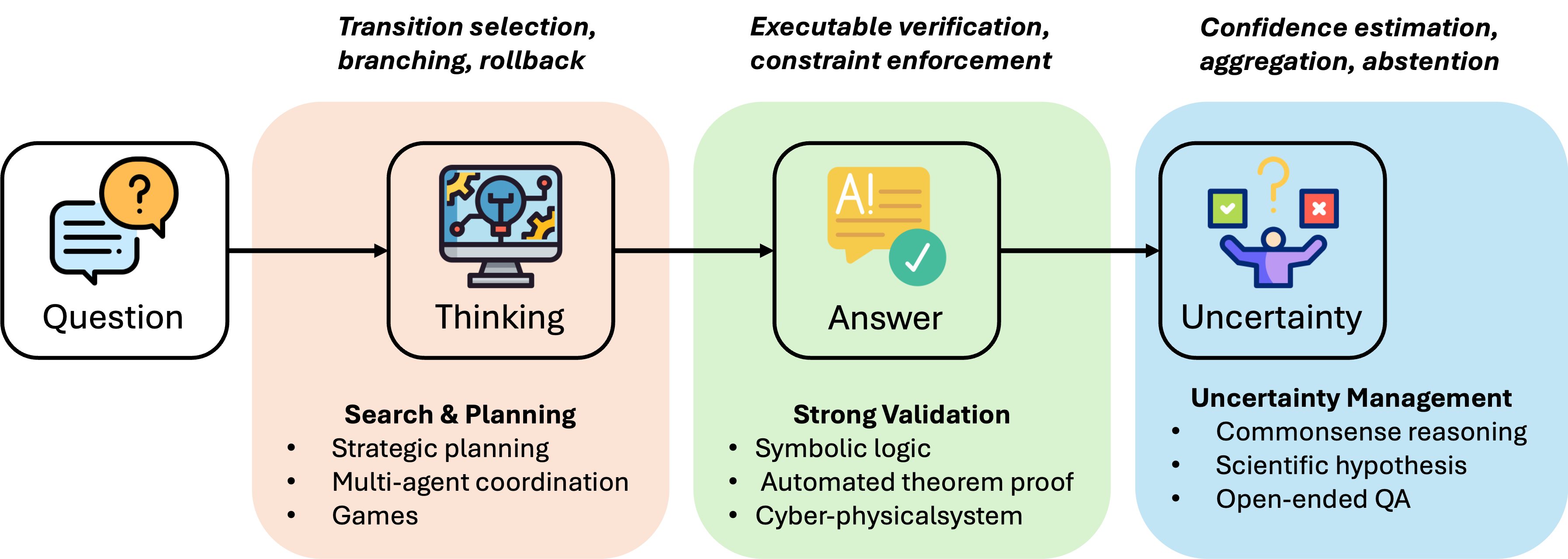}
    \caption{Domain-dependent control requirements. Different problem domains place their primary control burden on different components: search- and planning-intensive domains on transition structuring and search, validation-heavy domains on executable checking, and uncertainty-dominated domains on estimation and abstention. In statistical terms, these regimes emphasize the proposal kernel and posterior search, the likelihood or observation model, and, when ambiguity is intrinsic, the irreducible aleatoric component, respectively. This domain-level view complements the failure-level diagnostic in Table~\ref{tab:tab1}}
    \label{fig:fig4}
\end{figure}

We now survey methods according to the loop component they primarily implement. For each pattern we ask the same three questions: which statistical object it realizes (a belief statistic, a proposal kernel, a likelihood, a posterior-search procedure, an uncertainty estimate, or a decision rule), which error component it reduces (bias, variance, aleatoric or epistemic uncertainty, or state divergence), and which it cannot reduce, which is where a mismatch would arise. Sections~\ref{sub:state}--\ref{sub:uncertainty} cover the five core loop components, and Section~\ref{sec:crosscutting} covers the two cross-cutting dimensions of distributing and internalizing control. Because most methods realize several components at once, Table~\ref{tab:tab2} first maps representative methods onto the full set, and Appendix~\ref{app:mechanisms} gives a closer view of the most representative of these, the mechanism each uses and what it leaves uncontrolled; the per-component discussion that follows then examines each axis in depth. Different application domains stress different components, and we fold representative domains into the relevant component rather than treating applications separately. As shown in Figure~\ref{fig:fig4}, search- and planning-intensive domains stress transition structuring, branching, and rollback; validation-heavy domains such as symbolic logic and theorem proving stress executable checking; and uncertainty-dominated domains such as commonsense and open-ended reasoning stress estimation and risk-aware decision making.

For clarity, abbreviations used in Table~\ref{tab:tab2} include Program-Aided Language Models (PAL), Program-of-Thoughts (PoT), Reasoning and Acting (ReAct), and reinforcement learning from human feedback (RLHF).

\begin{table}[t]
\centering
\small
\caption{The diagnostic read against published results. Each row reports an existing finding, the failure component suggested by our framework, and why the intervention is consistent with that diagnosis. These studies do not causally identify the component itself. Numbers are quoted from the cited papers and are comparable within a row, not across rows.}
\label{tab:tab3}
\begin{tabularx}{\textwidth}{@{}p{4.6cm}p{2.5cm}X@{}}
\toprule
Published result & Dominant component & Reading under the decomposition \\
\midrule
Self-consistency: GSM8K $56.5\!\rightarrow\!74.4$ but CommonsenseQA $79.0\!\rightarrow\!80.7$ (PaLM-540B)~\citep{wang2022self} & \wt{Task-dependent benefit from sampling} & \wt{Aggregation helps substantially more on GSM8K in this setting, but these two results alone do not identify which error component caused the difference.} \\
GSM-HARD: chain-of-thought $23.1$ vs program-aided $61.2$ (Codex)~\citep{gao2023pal} & \wt{Executable computation} & \wt{Delegating arithmetic to an interpreter improves this large-number task; the comparison does not measure error correlation across samples.} \\
Longer reasoning or more inference-time compute leaves accuracy flat or lower~\citep{chen2025think23overthinkingo1like, wang2025thoughtsplaceunderthinkingo1like} & \wt{Possible effort mismatch} & \wt{The result is consistent with misallocated effort, but the cited studies do not isolate which control component should have received the budget.} \\
\bottomrule
\end{tabularx}
\end{table}

\subsection{Explicit State Representation} \label{sub:state}

The problems that place their primary control burden on state representation are those that must track many interacting quantities at once: multiple intermediate variables, evidence sources, dependencies among partial conclusions, or long-range context. In estimation terms the state is the statistic on which the downstream decision conditions, and the failure here is an \emph{insufficient statistic}: once instance-specific information needed for a later step has been dropped from the representation, no post-processing of that representation alone can recover it, by the data-processing inequality, except through priors or new observations such as retrieval, tools, or clarification. This is why the effective response is often to preserve and, where control requires it, externalize the relevant state rather than merely lengthen the reasoning trace, since adding more chain-of-thought to a problem whose needed intermediate information has already been lost does not recover it~\citep{nye2021show}.

Concretely, a system cannot constrain, validate, or correct a process whose intermediate state is inaccessible. In standard LLM generation, intermediate computation remains largely implicit in hidden activations, leaving little external surface for monitoring or intervention. Methods in this pattern address this foundational limitation by externalizing intermediate traces or working-state interfaces into observable representations, typically token sequences or structured objects, that can be inspected, compared, and reused. These interfaces need not be faithful readouts of the model's latent computation.

\begin{figure}
    \centering
    \includegraphics[width=.8\linewidth]{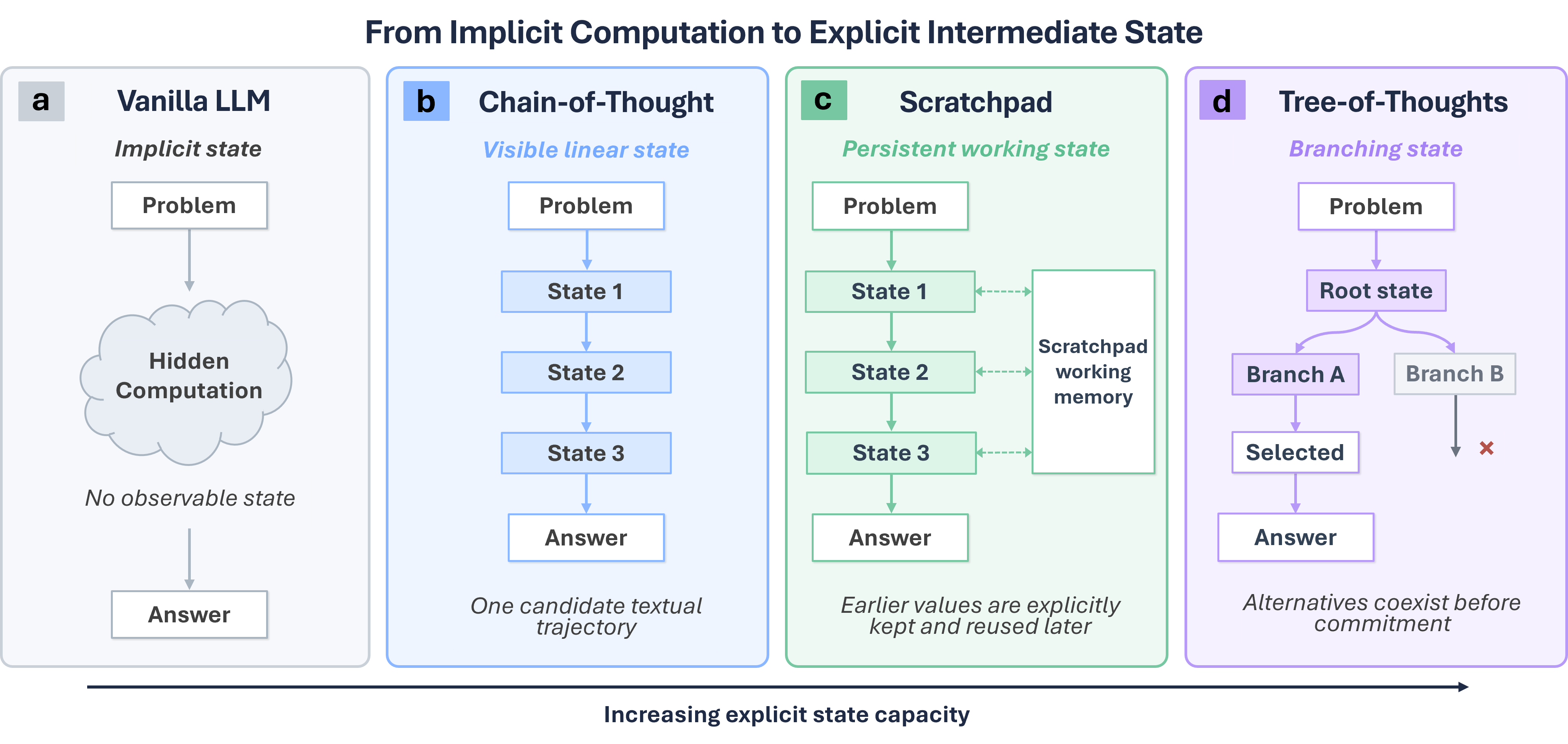}
    \caption{From implicit computation to explicit intermediate state.
\textbf{(a)} In vanilla LLM generation, intermediate computation remains hidden, so no observable state is available between the problem and the final answer.
\textbf{(b)} Chain-of-Thought prompting externalizes a single linear sequence of textual states, making the intermediate trajectory visible.
\textbf{(c)} Scratchpad prompting strengthens this representation by explicitly storing intermediate values in a working memory that can be reused across later steps, yielding persistent working state.
\textbf{(d)} Tree-of-Thoughts further generalizes explicit state into a branching structure, allowing multiple candidate states to coexist before commitment.
Across this progression, explicit state capacity increases from implicit internal computation to visible, persistent, and branching intermediate state; however, these methods primarily enrich state representation rather than directly constraining how state transitions are updated
}\label{fig:fig5}
\end{figure}

We organize these methods along three dimensions of increasing representational richness: \emph{visibility}, whether intermediate states are externalized at all; \emph{persistence}, whether earlier states remain accessible for later reuse; and \emph{structure}, whether multiple alternative states can coexist. Chain-of-Thought prompting provides linear textual visibility, Scratchpad prompting adds persistence through explicit working memory, and Tree-of-Thoughts prompting introduces branching state with parallel candidate trajectories. Figure~\ref{fig:fig5} illustrates this progression.

Importantly, an inspectable state interface is often useful for external process control but is not by itself sufficient for control. These methods make the reasoning trajectory observable, thereby enabling downstream control components such as validation (Section~\ref{sub:validation}) and rollback (Section~\ref{sub:search}), but they do not by themselves constrain transitions, verify intermediate claims, or support principled recovery from error. Moreover, the states they externalize are still predominantly free-form token sequences with weak typing, limited semantic annotation, and no dependency tracking. The limitations of these flat representations motivate the richer, structured state designs discussed in Section~\ref{sub:state-future}.

\subsubsection{Chain-of-Thought: Linear Textual State}

Chain-of-Thought (CoT) prompting externalizes intermediate reasoning by eliciting step-by-step traces as part of the model's output, providing a visible linear textual interface to otherwise latent computation~\citep{wei2022chain}. Such a trace is an observable representation available to the controller, not a guarantee that it faithfully exposes the model's latent computational state.
This addresses the most basic control deficit: when all reasoning remains implicit, there is no basis for inspection, monitoring, or intervention by downstream modules or human users. CoT thus provides a minimal externalized state interface without imposing additional representational structure.
Variants such as zero-shot CoT~\citep{kojima2022large}, automatic construction of CoT demonstrations~\citep{zhang2022automatic}, self-consistency decoding~\citep{wang2022self}, and prompt-adaptation methods~\citep{pitis2023boosted, wan2023better, yuan2024instance} reduce manual effort or improve robustness, but do not alter the underlying state representation, which remains a linear textual trajectory for each sampled reasoning path.

From a control perspective, however, CoT's contribution is confined to visibility alone. The trace is linear, with sequential commitment to each step and no built-in mechanism for revising earlier states or representing alternatives.
Intermediate values appear in the context window as part of the generated trace,
but they are not explicitly indexed, typed, or managed as reusable working state. The model has no dedicated mechanism to retrieve or update a specific earlier result.
Most importantly, transitions between steps are not explicitly governed by a dedicated
transition operator, validator, or rollback mechanism.
As a result, early errors can propagate through a trajectory, while fluent but invalid continuations may persist because CoT supplies no built-in validation or rollback mechanism~\citep{creswell2022faithfulreasoningusinglarge,turpin2023languagemodelsdontsay}.
In this sense, CoT makes reasoning legible, but leaves persistence (addressed by Scratchpad prompting), branching (addressed by Tree-of-Thoughts), validation and recovery (addressed in Sections~\ref{sub:search} and~\ref{sub:validation}) largely uncontrolled.

\subsubsection{Scratchpad: Persistent Working State}

Scratchpad prompting extends explicit state representation by treating the intermediate trace as a persistent working memory in which models generate, store, and reuse intermediate computations before producing a final answer. Originally proposed by \citet{nye2021show} through supervised training, scratchpads enable models to offload multi-step computation into observable tokens, improving performance on tasks requiring long or precise computations. Unlike CoT, where intermediate values appear sequentially but are not explicitly managed, the scratchpad provides a workspace in which earlier results are stored and reused by later steps, strengthening the persistence of intermediate state and increasing the surface available for external verification.

Subsequent work has extended the scratchpad concept along several axes. Self-notes~\citep{lanchantin2023learning} interleave reasoning tokens during input processing, creating persistence that is temporally distributed rather than deferred to the output stage. RAISE~\citep{liu2024llm} combines a transient scratchpad with retrieval-based long-term memory, extending persistence across multi-turn interactions. Inductive scratchpads~\citep{abbe2024far} encode task-specific structure into the workspace to improve length generalization,
while external interpreters and other tools can complement a scratchpad by computing or storing selected values, a capability that belongs to the broader tool-augmentation setting rather than to the scratchpad representation itself~\citep{mialon2023augmented,schick2023toolformer}.

Despite these advances, scratchpad methods often depend on careful prompt design or supervision~\citep{nye2021show, sahoo2024systematic} and the availability of an external trace does not by itself guarantee faithful reasoning~\citep{lanham2023measuring, turpin2023languagemodelsdontsay}; their token-based workspaces also grow with the computation and can become difficult to use reliably over long sequences~\citep{nye2021show,abbe2024far}. From a control perspective, scratchpads strengthen state persistence, which is the key limitation of linear CoT, but they do not explicitly constrain or validate transitions between states. The remaining gap in state representation is structural: neither CoT nor scratchpads maintain alternative candidate states, a capability introduced by Tree-of-Thoughts prompting.

\subsubsection{Tree-of-Thoughts: Branching State Structure}

Tree-of-Thoughts (ToT) prompting further generalizes explicit state representation by organizing intermediate reasoning into a branching structure in which multiple candidate states are maintained in parallel, enabling delayed commitment and limited backtracking~\citep{yao2023tree}. Within this paradigm, intermediate reasoning steps are organized as nodes in a tree, from which multiple candidate continuations can be generated and evaluated before the system commits to a single path forward.

Subsequent work has extended ToT along two axes. One line generalizes the tree topology itself: richer compositional variants~\citep{besta2024graph, zhang2023cumulative, saha2023branch} allow merging, accumulation, and reuse of intermediate states, while forest-based methods~\citep{bi2024forest} pursue broader parallel exploration across multiple trees.
A second line improves branch evaluation and search through planning-based algorithms~\citep{sel2023algorithm, hao2023reasoning}, skeleton-guided parallel expansion~\citep{ning2023skeleton}, uncertainty modeling~\citep{mo2024tree}, multi-agent coordination~\citep{haji2024improving}, or domain-specific adaptation~\citep{kim2023tree, yang2025qm}.

From a control perspective, the primary contribution of ToT-style methods lies in explicitly representing alternative intermediate states through branching. However, branch generation, scoring, and pruning typically rely on heuristic or model-based evaluations that may themselves be unreliable~\citep{yao2023tree, mo2024tree}.
Moreover, most ToT variants lack mechanisms for systematic learning from past failures or principled revision of evaluation criteria, limiting their effectiveness in long-horizon or highly uncertain settings~\citep{hui2024rot}. As a result, ToT enriches state representation through branching, while leaving transition validation and error correction largely uncontrolled.

Across all three approaches, Figure~\ref{fig:fig5} illustrates a consistent pattern: explicit state representation progresses from visibility through persistence to structure, but no method in this pattern constrains how states are updated, which is a gap addressed by the transition, search, and validation components developed in the following sections.

\subsection{Structured Transitions: Decomposition, Planning, and Program Induction} \label{sub:transition}

\begin{figure}
    \centering
    \includegraphics[width=.8\linewidth]{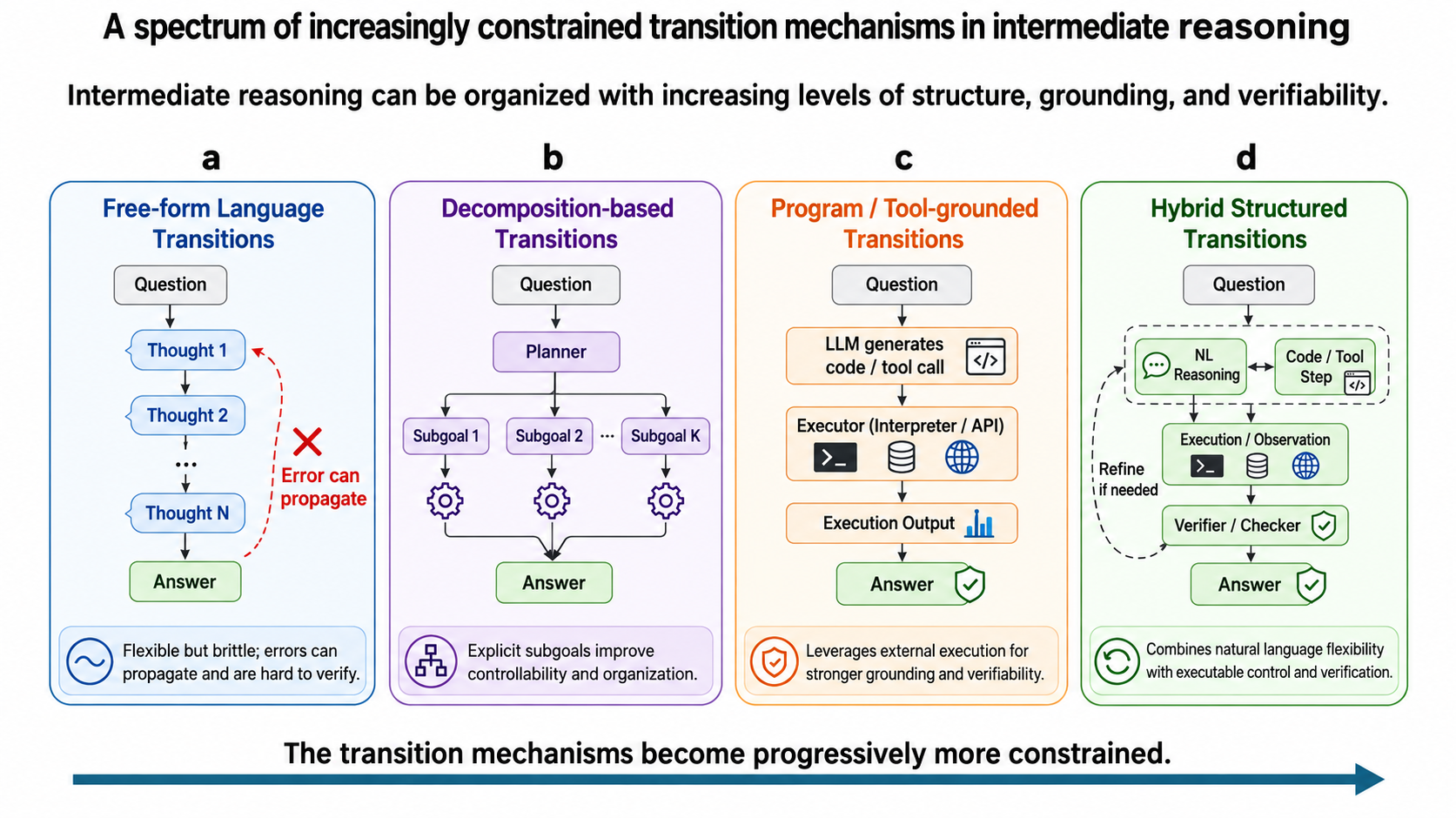}
    \caption{A spectrum of increasingly constrained transition mechanisms in intermediate reasoning.
\textbf{(a)} Free-form language transitions rely on unconstrained natural language to move between intermediate steps.
\textbf{(b)} Decomposition-based transitions introduce explicit plans, subgoals, or ordered subproblems to regulate how the reasoning process proceeds.
\textbf{(c)} Program- and tool-grounded transitions further constrain intermediate updates by translating reasoning into executable code or tool-invocation steps.
\textbf{(d)} Hybrid structured transitions combine natural language reasoning with executable programmatic steps, integrating the interpretability of textual explanation with the precision of symbolic execution
}\label{fig:fig6}
\end{figure}

Transition structuring concerns how the base proposal is shaped into the controlled transition kernel $q(s_{t+1}\mid s_t,a_t)$ of the filter, and it becomes the dominant bottleneck when failures originate in how the problem is decomposed or formulated: a wrong decomposition, a drifting output format, or a mismatch between planner and executor. A misspecified kernel can induce persistent bias across many of the trajectories it generates; additional sampling does not guarantee correction when the same flawed decomposition remains dominant. The intervention that helps is to constrain or repair the kernel itself rather than to add exploration on top of it~\citep{zhou2023leasttomostpromptingenablescomplex}. Figure~\ref{fig:fig6} places the mechanisms in this section on a spectrum from free-form to fully program-grounded transitions.
\subsubsection{Free-form Language Transitions}

In unconstrained prompting setups, LLMs often move between intermediate steps through free-form natural language generation. Such traces can be fluent and locally coherent, but they are fragile under small changes in the prompt, model, or context. Three recurring sources of brittleness are underspecification, error propagation, and limited grounding and verification.

A first source is underspecification. When intermediate requirements remain implicit, models may rely on defaults that are unstable across deployments. Prompt underspecification can induce behavioral drift, including regressions under model or prompt changes and failures even when additional requirements are provided \citep{yang2025promptsdontsayunderstanding}. Related evidence appears when context should resolve ambiguity but does not do so reliably. Even with visual context, vision-language models may still fail to disambiguate underspecified statements, suggesting that missing constraints can destabilize intermediate updates \citep{zhou2025focusunderspecification}. It also complicates evaluation, because incomplete or open-ended queries make judgments ill-posed and can shift preferences toward surface features such as style \citep{malaviya2025contextualizedevaluations}.

A second source is error propagation across multi-step traces. Intermediate natural language steps may introduce hallucinated premises or imprecise deductions, and early mistakes can shape later steps. Deductive verification studies of chain-of-thought show that intermediate reasoning can accumulate errors and hallucinations, motivating step-wise verification procedures that isolate the premises needed for each subproblem \citep{ling2023deductiveverificationcot}. Related work uses confidence signals and step-level supervision, showing that unverified transitions provide weak signals about where a trajectory becomes unreliable \citep{lu2024autopsvprocesssupervised}.

One way to make such transitions checkable is to train a step-level verifier over partial solution prefixes that estimates the probability that the current prefix leads to a correct answer, and to flag the step at which that estimate drops sharply~\citep{lu2024autopsvprocesssupervised}. Because this turns a free-form transition into a trained validation signal, we treat it with the validation component in Section~\ref{sub:validation} rather than here.

Compared with free-form transitions, more structured or tool-grounded intermediate states can narrow the space of plausible updates and provide externally checkable signals \citep{mialon2023augmented}. Prior work also distinguishes tool-augmented learning from settings that require sequential tool use \citep{wang2024empoweringtoollearning}. Although such designs do not eliminate failure, they shift brittleness toward more identifiable system errors, including incorrect tool selection, tool hallucination, and execution-blocking format issues \citep{winston2025taxonomyfailures}.

\subsubsection{Decomposition-based Transitions}

Planning-based prompting makes the intermediate plan explicit and uses it to regulate how the system moves from $s_t$ to $s_{t+1}$. Rather than relying on a single end-to-end generation, the model first proposes a plan or sequence of subgoals and then executes them in order. This converts a weakly constrained update into smaller plan-conditioned updates, often instantiated as checklists, subquestions, or structured task graphs, with the goal of improving controllability \citep{wang2023planandsolvepromptingimprovingzeroshot,sun2024pearlprompting,zhou2023leasttomostpromptingenablescomplex}.

A first pattern is prompt-only planning, where planning and execution are both carried out in language. Plan-and-Solve asks the model to devise a plan and then solve accordingly, with evidence that explicit planning reduces missing-step errors in zero-shot reasoning \citep{wang2023planandsolvepromptingimprovingzeroshot}. PEARL similarly stages long-context reasoning through action mining, plan formulation, and plan execution \citep{sun2024pearlprompting}. Least-to-Most uses decomposition followed by sequential subproblem solving, where earlier answers support later steps, and is reported to improve easy-to-hard generalization by controlling the order and granularity of transitions \citep{zhou2023leasttomostpromptingenablescomplex}.

A second pattern separates planning from grounded execution in planner--executor architectures. ReWOO first constructs a plan with placeholders for evidence, then invokes workers or tools to fill those placeholders, and finally synthesizes an answer from the resulting evidence. This reduces redundancy from interleaving reasoning and observations, while yielding token savings and competitive accuracy on multi-step benchmarks \citep{xu2023rewoodecouplingreasoningobservations}. Related planner--executor patterns also appear in applied systems such as scientific workflows and interactive visual analytics \citep{xiao2024cellagentllmdrivenmultiagentframework,zhao2025lightva}. In these settings, the plan acts as an intermediate representation that constrains later actions and state updates.

Decoupling planning from evidence collection can also curb repeated prompt growth: ReWOO reuses plan and evidence placeholders instead of interleaving every reasoning step with every observation, and reports token-efficiency gains without claiming a general asymptotic bound~\citep{xu2023rewoodecouplingreasoningobservations}.

Planning narrows the space of transitions, but it does not guarantee correctness. If a plan is wrong, later updates may remain locally coherent while being globally misdirected, especially in long-horizon settings. One issue is that out-of-the-box LLMs are not reliably trained to produce accurate, executable plans for complex environments \citep{erdogan2025planandactimprovingplanningagents}. Moreover, when execution reveals mismatches between assumptions and reality, systems that simply follow the initial plan may continue along an erroneous trajectory. Recent frameworks therefore emphasize dynamic replanning and failure recovery. PLAN-AND-ACT highlights replanning as a key ingredient for robust web navigation \citep{erdogan2025planandactimprovingplanningagents}. InstructFlow further argues that naive retries or superficial replanning often repeat the same mistakes, and instead proposes failure-driven constraint induction for targeted repair \citep{chi2025instructflow}. These observations motivate search, branching, and rollback mechanisms for revising trajectories when intermediate evidence contradicts the current plan, which we discuss in the next subsection.

\subsubsection{Program- and Tool-Grounded Transitions}

Program-Aided Language models (PAL) translate a natural-language problem into an executable program and delegate the actual computation to an interpreter~\citep{gao2023pal}. This grounds transitions in an externally checkable substrate and addresses a specific failure of CoT-style reasoning, in which a fluent natural-language derivation can still yield an arithmetically wrong answer. Given a faithful problem-to-program translation and a reliable interpreter, program execution is deterministic; PAL therefore shifts arithmetic execution away from the language model but does not guarantee that the program correctly formulates the problem.

Beyond arithmetic accuracy, program-aided reasoning has been studied for whether execution-based signals improve calibration and selective prediction~\citep{kabra2023}; program grounding can also be paired with decomposition strategies such as Least-to-Most~\citep{zhou2023leasttomostpromptingenablescomplex}. What program grounding does not control is the correctness of the formulation itself: an interpreter faithfully executes a wrong program, so errors in translating the problem into code pass through unchecked, which motivates pairing program grounding with the validation component of Section~\ref{sub:validation}.

\subsubsection{Hybrid Structured Transitions}

Hybrid structured transitions split the proposal kernel into a semantic component handled by language and an exact component handled by execution, so that formulation stays flexible while computation is made deterministic. This targets a specific failure of monolingual CoT: when a single kernel must both formulate and compute, arithmetic errors and symbolic slips enter as noise in the transition even when the formulation is correct. Earlier approaches instead relied on large volumes of expert-annotated reasoning steps \citep{ling2017, cobbe2021training}, which are data-intensive and hard to scale, while few-shot scratchpads externalized intermediate computation~\citep{nye2021show}, and few-shot CoT subsequently elicited natural-language reasoning traces without parameter updates~\citep{wei2022chain}. CoT improves many reasoning tasks but still leaves arithmetic and symbolic execution to the model, motivating program-grounded alternatives~\citep{gao2023pal,chen2023pot}.

To address these limitations, the Program-of-Thoughts (PoT) framework expresses reasoning steps as executable programs and delegates their execution to an external interpreter, so that iterative computation and symbolic manipulation are handled by code rather than by the model itself~\citep{chen2023pot}. This isolates the part of the transition that must be exact, and the reported gains over CoT on math word problems and financial reasoning come from removing arithmetic and symbolic-manipulation errors rather than from more elaborate natural-language reasoning~\citep{chen2023pot}. Sampling, general tool use, critique, and self-evaluation are complementary mechanisms studied in separate frameworks~\citep{wang2022self,schick2023toolformer,gou2024critic,xie2023selfevaluation}; they should not be treated as PoT extensions without a direct integration study. Like PAL, PoT guarantees only the behavior of the generated program under the chosen interpreter; formulation errors and the security of executing generated code remain outside that guarantee.

What transition structuring leaves uncontrolled is the correctness of the structure itself. A well-formed plan or a syntactically valid program can still encode a wrong decomposition or a mistaken formulation, and constraining how one step follows another does not check whether that step is admissible. Transition structuring therefore composes with, but does not substitute for, the search and validation components developed in the following two subsections.

\subsection{Search, Rollback, and Iterative Correction} \label{sub:search}

\begin{figure}
    \centering
    \includegraphics[width=1\linewidth]{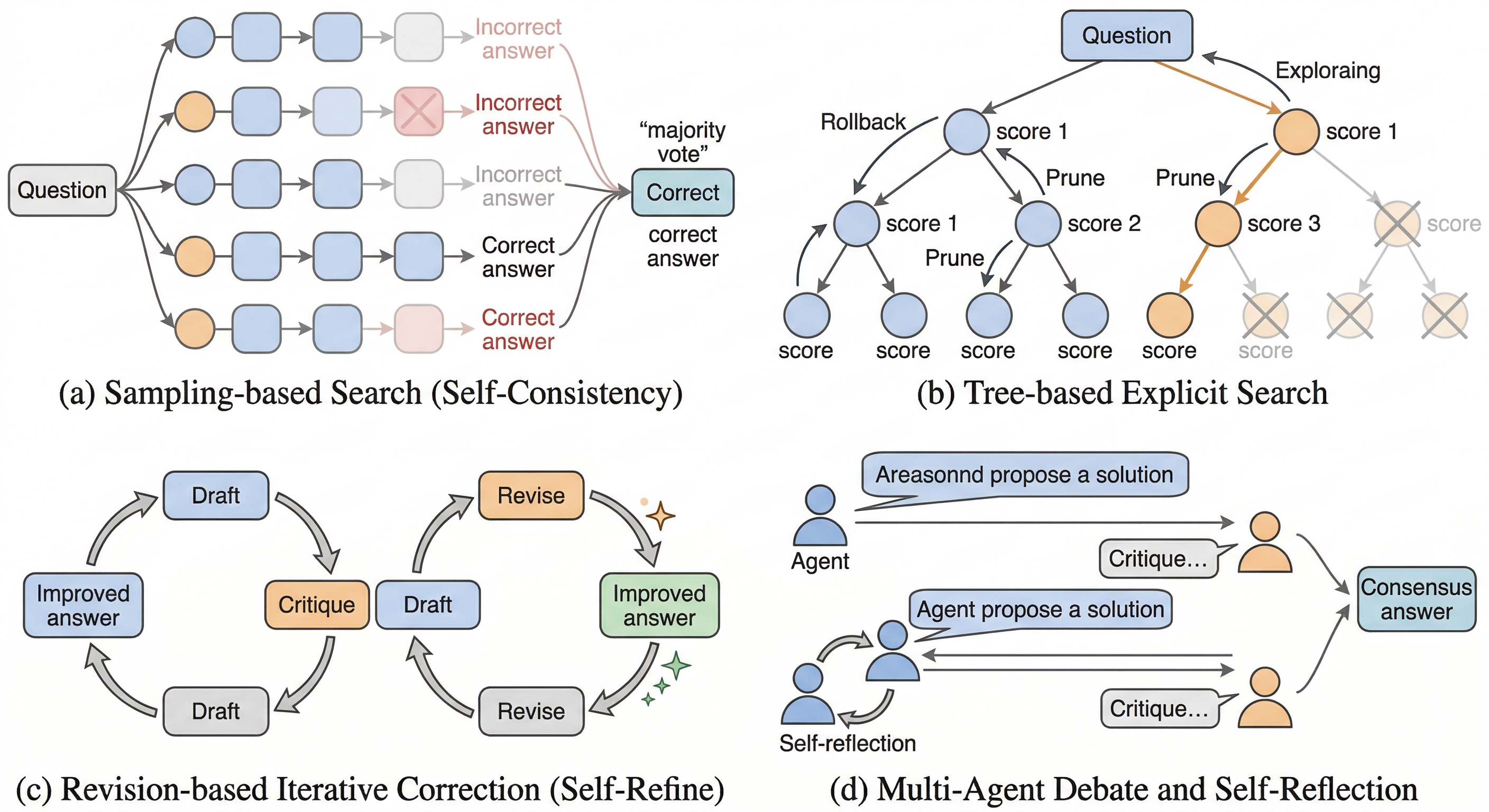}
    \caption{LLM Reasoning with Search, Rollback, and Iterative Correction. \textbf{(a)} Self-Consistency: Majority vote over sampled paths. \textbf{(b)} Tree Search: Systematic navigation with scoring and pruning. \textbf{(c)} Self-Refine: Iterative draft-critique-revise cycles. \textbf{(d)} Multi-Agent Debate: Interactive deliberation with critique and reflection to achieve consensus}
    \label{fig:fig7}
\end{figure}

Search and rollback matter most when the posterior over solutions is multimodal, so that several plausible paths compete and an early commitment collapses the belief onto one mode before the evidence justifies it. The bottleneck here is premature commitment, and validating a single trajectory is not enough; the system needs the ability to keep competing modes alive, compare them, and back out of a mode that later evidence contradicts. Conversely, when the posterior is essentially unimodal and the trajectory is forced, branching and rollback buy little and mainly increase cost~\citep{yao2023tree}. Figure~\ref{fig:fig7} illustrates the main families of search and correction covered below.

From the estimation-and-decision view of this survey, search, rollback, and iterative correction are not merely ways to produce longer reasoning traces. Rather, they are mechanisms for regulating how a solution trajectory is explored, evaluated, revised, and, when necessary, abandoned.

To make this idea precise, let \(x\) denote the input problem, and let \(s_t \in \mathcal{S}\) denote the intermediate state of the reasoning process at step \(t\). Here, a ``state'' may be a textual partial solution, a structured decomposition, a program fragment, or any other explicit intermediate representation maintained by the system. Let \(a_t \in \mathcal{A}(s_t)\) denote a candidate transition proposed from state \(s_t\), such as a next reasoning step, a subgoal, or a revision action. Executing that transition yields the next state
\[
s_{t+1} = f(s_t, a_t),
\]
where \(f\) denotes the induced state-update operator. A complete reasoning process can then be represented as a trajectory
\[
\tau = (s_0, a_0, s_1, a_1, \dots, s_T).
\]
Under this notation, the key control question is not only how to generate candidate transitions, but how to regulate exploration over alternative trajectories, decide when to prune or backtrack, and revise a partial trajectory when intermediate evidence indicates failure or inconsistency. This view is consistent with recent methods that treat LLM inference as deliberate search over intermediate thoughts, iterative self-repair, or feedback-guided correction rather than single-pass generation \citep{wang2022self,yao2023tree,madaan2023self,shinn2023reflexion}.

We group representative methods in this pattern according to the main control component they introduce. Sampling-based methods explore multiple complete trajectories in parallel and aggregate across them before committing to an answer. Tree-based methods maintain an explicit frontier of partial trajectories and selectively expand, prune, or revisit branches. Revision-based methods focus on local repair of an evolving draft rather than broad branching. Verifier-triggered methods strengthen correction by using external or semi-external feedback signals to determine when rollback or revision is needed. Although these methods differ in implementation, they share the same objective: to improve trajectory-level robustness under uncertainty.

\subsubsection{Sampling-based Search}

Sampling-based methods regulate commitment by generating multiple candidate trajectories from the same initial problem state and aggregating their outcomes. Instead of trusting a single sampled reasoning chain, the controller allocates computation across repeated rollouts and then applies a selection rule, such as majority voting or verifier-based reranking, to determine the final answer. The core control component is therefore parallel exploration followed by aggregation.

Formally, suppose the model samples \(K\) candidate trajectories
\[
\tau^{(k)} \sim p_\theta(\tau \mid x), \qquad k = 1, \dots, K,
\]
where \(p_\theta(\tau \mid x)\) denotes the model-induced distribution over reasoning trajectories given input \(x\). Let \(y^{(k)} = g(\tau^{(k)})\) denote the final answer extracted from the \(k\)-th trajectory. The simplest aggregation rule is majority voting,
\[
\hat y
=
\arg\max_y
\sum_{k=1}^K \mathbf{1}\{y^{(k)} = y\},
\]
which selects the answer that appears most often among the sampled trajectories. In other words, the controller does not ask which single trajectory looks best in isolation, but which answer is most stable across diverse rollouts.

A representative example is self-consistency, which replaces greedy chain-of-thought decoding with sampling over multiple reasoning paths and selects the most consistent answer by marginalizing over them \citep{wang2022self}. The key intuition is that a complex reasoning problem may admit several distinct valid paths even when it has a unique correct answer. By aggregating across trajectories, self-consistency reduces the influence of incidental local errors in any single rollout and improves performance on arithmetic and commonsense reasoning tasks \citep{wang2022self}. More broadly, repeated sampling can be viewed as an inference-time scaling strategy, in which additional computation is spent on generating and comparing multiple candidate trajectories rather than committing immediately to one path \citep{brown2024large}.

Read as an estimator, self-consistency is a Monte Carlo estimate of the answer marginal $p(y\mid x)=\int \mathbf{1}\{g(\tau)=y\}\,p_\theta(\tau\mid x)\,d\tau$, and its error is governed by the variance of that estimate. For a scalar statistic averaged over $K$ exchangeable samples with common pairwise correlation $\bar\rho$, the standard effective-sample-size approximation is
\begin{equation}
K_{\mathrm{eff}} = \frac{K}{1 + (K-1)\bar\rho},
\label{eq:ess}
\end{equation}
so the variance of that sample mean falls like $1/K_{\mathrm{eff}}$ rather than $1/K$. This formula is only a heuristic for nonlinear majority voting, but it captures the loss of effective diversity as samples become correlated. Shared misconceptions can therefore keep effective diversity low, and additional samples provide no guarantee of correcting a dominant systematic error. This limitation recurs when we read the same method as an uncertainty estimator in Section~\ref{sub:uncertainty}.

A simple example helps clarify the control role. Suppose a model is solving a grade-school arithmetic problem. One sampled trajectory may make an early subtraction mistake, another may phrase the reasoning differently but compute correctly, and a third may reach the same correct answer through a different decomposition of the quantities involved. Sampling-based search delays commitment until these trajectories can be compared at the level of final answers, so the controller is less dependent on any one flawed path.

From a control perspective, the main benefit of this pattern is robustness to stochastic local failures. It is particularly effective when the model can produce diverse but partially redundant reasoning paths, so that the correct answer recurs more reliably than any individual mistake. However, what remains largely uncontrolled is targeted recovery. These methods typically do not maintain an explicit notion of rollback to a particular earlier state, nor do they identify which intermediate step caused an error. If the model's sampled trajectories are highly correlated, or if they all reflect the same underlying misconception, aggregation may simply reinforce a shared mistake. Moreover, the computational cost grows directly with the number of sampled rollouts \citep{wang2022self,brown2024large}. Thus, sampling-based search delays commitment, but it does not by itself provide localized revision or principled backtracking.

\subsubsection{Tree-based Explicit Search}

Tree-based search methods make branching explicit and treat reasoning as controlled exploration over partial trajectories. Instead of sampling several complete solutions independently, these approaches maintain a frontier of intermediate states, expand selected nodes, evaluate candidate continuations, and prune or revisit branches when a path appears unpromising. The control component is therefore explicit search over partial states with selective expansion and rollback.

Let
\[
\mathcal{F}_t = \{s_t^{(1)}, \dots, s_t^{(m_t)}\}
\]
denote the frontier of active partial states at search depth \(t\). For each frontier state \(s \in \mathcal{F}_t\), the model proposes candidate transitions \(a \sim p_\theta(a \mid s)\), producing child states \(s' = f(s,a)\). A scoring function \(V(s') \in \mathbb{R}\) then estimates how promising each partial state is. The controller keeps only the most promising candidates, for example through a rule of the form
\[
\mathcal{F}_{t+1}
=
\operatorname{TopB}\bigl(\{\,s' : s' \in \mathrm{Expand}(\mathcal{F}_t)\,\}; V \bigr),
\]
where \(\operatorname{TopB}\) retains the top \(B\) states according to the score \(V\). In this way, the system regulates not only which answer to output, but which partial trajectories deserve further computation.

Tree-of-Thoughts is the clearest example of this paradigm \citep{yao2023tree}. It generalizes linear chain-of-thought prompting by organizing reasoning into coherent thought units and interleaving two control operations: generation of candidate next thoughts and evaluation of which thoughts to expand \citep{yao2023tree}. The framework is designed specifically to support exploration, strategic lookahead, and backtracking when early decisions matter for global success. Related extensions broaden the search space beyond trees. Graph-of-Thoughts allows intermediate states to be merged, refined, or recombined in a more general graph structure \citep{besta2024graph}, while Monte Carlo tree search (MCTS)-style approaches use rollout-based evaluation and search policies to balance exploration and exploitation during reasoning \citep{zhang2024rest,park2025ensembling,gao2024interpretable,xie2024monte}.

A simple planning example illustrates why this matters. Suppose the system is solving a problem that requires decomposing a task into subgoals. One branch may begin with a decomposition that looks plausible locally but later makes the task infeasible. Another may appear less direct at first but remains globally consistent. A tree-based controller can preserve both branches in parallel, expand them selectively, and return to an earlier branch if later evidence suggests that the current one is failing. This is qualitatively different from sampling-based aggregation, which compares complete trajectories only after they are finished.

From a control perspective, the main advantage of tree-based search is that it regulates commitment at the level of intermediate states rather than only final answers. This is especially useful in long-horizon tasks, combinatorial reasoning, planning, and settings where an early local decision can constrain whether later recovery is possible. Compared with pure repeated sampling, explicit search can allocate computation more selectively by concentrating expansion on promising branches. It also externalizes control decisions, such as which nodes were expanded or pruned, making the search process more interpretable.

Its main limitation is that the quality of control depends heavily on the reliability of the scoring function \(V\). If branch evaluation is weak or miscalibrated, valid trajectories may be pruned too early, while invalid ones may be explored extensively. In addition, search cost grows rapidly with depth and branching factor, so practical systems often rely on heuristics such as beam width limits, shallow lookahead, or approximate self-evaluation \citep{yao2023tree,zhang2024rest,park2025ensembling}. These heuristics can themselves reintroduce premature commitment. Recent analyses further show that verifier-guided search may fail when evaluator errors compound, so stronger search requires not only broader exploration but also reliable intermediate scoring \citep{yu2025scaling}. Thus, explicit search improves flexibility and observability, but does not by itself guarantee stable recovery from error.

\subsubsection{Revision-based Iterative Correction}

Revision-based methods emphasize local repair rather than broad exploration. Instead of maintaining many competing branches, they begin with a draft trajectory, generate feedback on its weaknesses, revise the draft accordingly, and repeat this process until a stopping criterion is met. The control component is therefore iterative local correction of an evolving solution.

Let \(z^{(0)}\) denote the initial draft generated for input \(x\). At refinement round \(r\), a feedback operator produces critique
\[
c^{(r)} = C_\theta(x, z^{(r)}),
\]
and a revision operator uses that critique to produce an updated draft
\[
z^{(r+1)} = R_\theta(x, z^{(r)}, c^{(r)}).
\]
Here \(z^{(r)}\) is the current working state of the solution, \(c^{(r)}\) is feedback describing what should be improved, and \(R_\theta\) is the refinement step that updates the draft. Unlike tree-based search, which keeps multiple partial trajectories alive, revision-based correction focuses computation on repairing the current one.

Self-Refine is the canonical example of this approach. It alternates between feedback and refinement, using the same model to generate an initial output, critique it, and then improve it iteratively, without additional training or external supervision \citep{madaan2023self}. Reflexion extends this idea by storing reflective feedback in an episodic memory buffer, so that information from previous failed attempts can influence later ones \citep{shinn2023reflexion}. In that case, one may write an additional memory update
\[
m^{(r+1)} = U(m^{(r)}, c^{(r)}),
\]
where \(m^{(r)}\) denotes the accumulated reflective memory after round \(r\). The next refinement can then depend on both the current draft and the stored memory. This makes the repair process persistent across trials rather than confined to a single draft \citep{shinn2023reflexion}.

A concrete example is useful here. Suppose a draft solution to a math or logic problem contains a missing constraint, such as failing to account for an excluded case. A critique step may identify that omission explicitly, and the next refinement step can revise the affected portion of the reasoning without regenerating the entire solution from scratch. In this sense, revision-based methods act more like local debugging than global search.

From a control perspective, this pattern is attractive because it offers a lightweight recovery mechanism when errors are local, visible, and diagnosable from the current draft. It can improve coherence, clarity, and correctness without the computational cost of maintaining many parallel branches. However, what remains uncontrolled is reliable diagnosis. If the critique signal is weak, incomplete, or biased by the same misconception that generated the original draft, revision may simply rephrase an error rather than correct it. An incorrect latent assumption can survive multiple rounds of refinement even while the surface form improves. More iterations also increase latency and may degrade into repeated rewriting without meaningful state revision \citep{madaan2023self,pan2023automatically}. Thus, iterative correction is often strongest when paired with a more grounded feedback signal than self-critique alone.

\subsubsection{Verifier-triggered Correction Loops}

Verifier-triggered correction loops strengthen revision by introducing an external or semi-external signal that determines when correction is needed and, in some cases, where it should be applied. The control component is feedback-conditioned rollback: after generating a candidate trajectory, the system checks it against a verifier and then revises, resamples, or backtracks if the trajectory fails that check. In this subsection, verification is relevant specifically as a trigger for recovery, rather than as a broader validation framework.

At the trajectory level, let \(v(\tau)\) denote a verifier score for a complete candidate trajectory \(\tau\). If the system has generated multiple candidates \(\tau^{(1)},\dots,\tau^{(K)}\), it may select
\[
\hat{\tau} =
\arg\max_{\tau\in\{\tau^{(1)},\dots,\tau^{(K)}\}} v(\tau).
\]
This formulation captures verifier-based reranking, where the generator proposes candidate solutions and a separate verifier judges which one is most likely to be correct. Training verifiers for math word problems uses exactly this pattern: multiple candidate solutions are generated, and a verifier is trained to rank them, improving performance on GSM8K \citep{cobbe2021training}.

More fine-grained control is possible at the step level. Let
\[
v_t(s_t, a_t, s_{t+1})
\]
denote a local verifier score for the transition from \(s_t\) to \(s_{t+1}\) via action \(a_t\). Then a low verifier score can trigger rollback:
\[
\text{if } v_t(s_t, a_t, s_{t+1}) < \gamma,
\quad \text{rollback and regenerate from an earlier state.}
\]
Here, \(\gamma\) is a threshold below which the current transition is judged unreliable. This step-level view is closely aligned with work on step-aware verifiers and process supervision, where intermediate reasoning steps are evaluated directly rather than only the final answer \citep{li2022making,lightman2023let,wang2024mathshepherd}.

Chain-of-Verification provides another useful example \citep{dhuliawala2024chain}. Instead of simply asking the model to critique its own answer in free form, it introduces a structured procedure: draft an initial response, plan verification questions, answer those questions independently, and then generate a final revised response. In notation, one may write
\[
d = G_\theta(x), \qquad
q_{1:m} = P_\theta(x,d), \qquad
a_i = A_\theta(x,q_i), \qquad
\hat y = U_\theta(x,d,\{a_i\}_{i=1}^m),
\]
where \(d\) is the initial draft, \(q_{1:m}\) are verification questions, \(a_i\) are independently generated answers to those questions, and \(\hat y\) is the final updated response. The key control idea is that the feedback used for revision is made more diagnostic and less entangled with the original draft.

A simple code-generation example illustrates the advantage of verifier-triggered correction. If a model proposes a program that fails to compile or does not pass its unit tests, those failures provide concrete evidence that the current trajectory is invalid. The controller can then revise the code, resample a candidate, or roll back to an earlier partial solution. Such feedback is more directly grounded in task behavior than free-form self-critique, although incomplete tests still provide only partial evidence~\citep{chen2024teaching}.

From a control perspective, the main benefit of verifier-triggered correction is that it provides a more reliable feedback channel for recovery. When the verifier is grounded and diagnostic, the system is less likely to overlook errors that would persist through fluent but unsupported revisions. This makes such methods especially useful in domains such as code generation, mathematical reasoning, and formal tasks where at least partial checking is available. However, verification does not automatically solve recovery. Many tasks do not provide complete oracles, and even strong verifiers may be sparse, delayed, or imperfect. A failed check indicates that the current trajectory is inadequate, but it does not by itself determine how far the system should roll back, which part of the trajectory caused the failure, or what alternative should be tried next. Thus, adding a verifier improves the feedback available for control, but robust correction still depends on effective policies for localization, selection, and rollback \citep{pan2023automatically,yu2025scaling}.


\noindent\textbf{Representative domains.}
Search- and planning-intensive domains make the search and rollback components explicit. In competitive programming, AlphaCode generates a large pool of candidate programs and then filters and clusters them, an explicit generate-and-test search over solutions rather than a single committed trajectory~\citep{li2022competition}. In sequential decision making, Reasoning via Planning couples an LLM world model with Monte Carlo Tree Search, so that candidate steps are expanded, scored, and revised under lookahead instead of being committed greedily~\citep{hao2023reasoning}. The same pattern appears in embodied and cyber-physical planning, where systems such as SayCan separate proposal from feasibility by scoring candidate skill sequences against grounded value estimates before execution~\citep{ahn2022can}. In each case the control content lies not in the fluency of any single plan but in the ability to explore alternatives and abandon unpromising branches.

\subsection{Validation and Constraint Enforcement} \label{sub:validation}

Validation is the action-conditioned observation model $p(o_{t+1}\mid s_{t+1},a_t)$ of the filter, the channel through which evidence made available by a control action reweights the belief, and it is the binding constraint whenever correctness is decided by something external to fluent text: hard constraints, unit tests, symbolic soundness, or factual grounding. When such a channel is missing, longer explanations may increase apparent plausibility without supplying independent evidence of correctness; where executable or symbolic checks exist, they can directly test properties that further deliberation alone cannot~\citep{chen2021evaluating}. The reliability of validation is exactly the question of whether this observation model is well specified: a miscalibrated or gameable verifier is a misspecified likelihood that can move the belief in the wrong direction.

Validation and constraint enforcement regulate whether a candidate state or transition is admissible before the reasoning trajectory is allowed to continue. From the estimation-and-decision view of this survey, these methods do not primarily improve CPS by producing longer or more fluent reasoning traces; they improve it by supplying rejection signals, executable checks, or action-space restrictions that prevent locally plausible but globally invalid trajectories from propagating. Let \(s_t\) denote the current state, \(a_t\) a proposed transition, and \(s_{t+1}\) the resulting candidate state. A generic validator can be written as
\[
z_t = V(s_t, a_t, s_{t+1}) \in [0,1],
\qquad
\text{accept if } z_t \ge \tau_v,
\]
where \(V\) may be realized by the model itself, by an external executor, by decode-time constraints, or by a pipeline that combines several such signals. The central design question is therefore not whether a system can propose intermediate reasoning steps, but how reliably it can reject, revise, or block those steps when they violate constraints or exceed the current validation budget.

\subsubsection{Model-internal Validation}
\label{sec:model-internal-validation}

The cheapest validation component available to a control-centric reasoner is the
generator itself: rather than committing to the first emitted trajectory, the
model is asked to re-examine its intermediate states and reject those it judges
inconsistent. Early instantiations of this idea include backward
self-verification, where conclusions reached by chain-of-thought are fed back
as conditions to recompute the original premises and discrepancies are used as
a correctness signal \citep{weng2023large}; iterative self-feedback frameworks
such as \textsc{Self-Refine}, in which the same LLM generates, critiques, and
revises its own output until a stopping criterion is met \citep{madaan2023self};
and Reflexion, which converts environment or simulated feedback into verbal
self-reflections stored in episodic memory to bias subsequent rollouts
\citep{shinn2023reflexion}. Self-consistency, although introduced as a decoding
strategy, can also be read as a model-internal validator: by marginalising over
sampled trajectories and selecting the modal answer, it uses agreement among
internally generated paths as a proxy for correctness
\citep{wang2022self}. Chain-of-Verification extends this idea by
having the model plan and answer atomic verification questions about its own
draft before producing a final response, decoupling fact-checking from the
biases of the original generation context \citep{dhuliawala2024chain}.

A second pattern integrates validation directly into the search process.
Self-Evaluation Guided Beam Search treats stepwise self-assessment as a
model-derived heuristic over partial trajectories, pruning branches the model
deems low-confidence \citep{xie2023selfevaluation}, while LLM-as-a-judge
protocols evaluate complete responses through pairwise or rubric-based prompts
and have become a standard scalable proxy for human preferences
\citep{zheng2023judging}. When step-level supervision is available, learned
process reward models (PRMs) generalise self-evaluation into a trained
verifier: \citet{lightman2023let} show that PRMs trained on dense human
annotations of intermediate steps substantially outperform outcome-only
verifiers on MATH,  and Math-Shepherd demonstrates that the same signal can be
distilled automatically by Monte-Carlo rollouts, eliminating per-step human
labelling \citep{wang2024mathshepherd}.

Within the control-centric framing, model-internal validation acts as a low-cost
but potentially miscalibrated feedback channel on trajectory execution.
Several recent studies document this limitation. Across the tested prompts
and tasks, \citet{huang2024selfcorrect} find that intrinsic self-correction can
degrade reasoning by changing initially correct answers. \citet{stechly2024self}
report collapse in several tested planning domains but a modest intrinsic gain
on Blocksworld; in every setting listed in Table~\ref{tab:tab4}, feedback
from a sound first-error verifier performs better. \citet{tyen2024llms} further
identify mistake localisation as a major bottleneck. These findings show
that a model-generated verification signal can share blind spots with the
trajectory it evaluates; they do not imply that every internal validator fails
in every setting.

Table~\ref{tab:tab4} collects representative numbers behind this reading. Across these arithmetic, commonsense, planning, and code settings, intrinsic self-correction is mixed and often negative, whereas the external or stronger feedback condition improves over the corresponding baseline in every listed row. The comparison supports treating validation as a distinct control component, while remaining specific to the models, prompts, tasks, and feedback sources reported by the cited studies.

\begin{table}[t]
\centering
\small
\caption{Intrinsic self-correction versus external verification, reported as accuracy or task success (\%). The intrinsic column shows performance after self-correction with no external signal; the external column shows performance when a sound verifier, an oracle stopping rule, or a stronger feedback source is available. Settings and metrics differ across papers, so entries are comparable within a row, not across rows.}
\label{tab:tab4}
\begin{tabular}{@{}llll@{}}
\toprule
Benchmark (model) & Intrinsic self-correction & External verification & Source \\
\midrule
GSM8K (GPT-4) & $95.5 \to 89.0$ & $95.5 \to 97.5$ & \wt{\citet{huang2024selfcorrect}} \\
\wt{CommonsenseQA} (GPT-3.5) & $75.8 \to 41.8$ & $75.8 \to 89.7$ & \wt{\citet{huang2024selfcorrect}} \\
Graph colouring (GPT-4) & $16 \to 2$ & $16 \to 38$ & \citet{stechly2024self} \\
Blocksworld (GPT-4) & $40 \to 55$ & $40 \to 87$ & \citet{stechly2024self} \\
Code repair, APPS (GPT-4) & \wt{$33.3$ (GPT-4 feedback)} & $52.6$ (human feedback) & \citet{olausson2024selfrepair} \\
\bottomrule
\end{tabular}
\end{table}

\subsubsection{External Executable Validation}
\label{sec:external-validation}

When the trajectory inhabits a domain admitting a deterministic executor, the
controller can replace self-judgment with task-grounded feedback for the
operations covered by that executor. The canonical instance is code execution:
\textsc{Self-Debug} teaches LLMs to inspect interpreter output and
\emph{rubber-duck} their generated programs through few-shot demonstrations,
turning runtime traces into a feedback signal that drives iterative repair
\citep{chen2024teaching}. CRITIC generalises this pattern by routing model
outputs through external tools, code interpreters, search engines,
fact-checkers, and conditioning the next-step generation on the resulting
critiques, with the authors arguing that reliable improvement is largely
\emph{contingent} on tool feedback rather than self-criticism
\citep{gou2024critic}. In code generation, unit tests provide a particularly
sharp verifier: HumanEval popularized pass@$k$ evaluation of functional
correctness through held-out unit tests~\citep{chen2021evaluating}, and
\citet{olausson2024selfrepair} show that the apparent gains of self-repair
collapse once one accounts for the inference budget, with most of the lift
attributable to the tests themselves rather than to the model's debugging
prose.

A parallel line offloads reasoning to symbolic systems. Logic-LM translates
natural-language premises into first-order logic, dispatches the formulation
to a satisfiability modulo theories (SMT) or theorem-proving backend, and only consults the LLM for parsing
and self-refinement against solver error messages \citep{pan2023logiclm}; LINC
adopts the same neuro-symbolic split and uses a first-order logic prover to
filter syntactically and semantically invalid translations before majority
voting \citep{olausson2023linc}. For numerical reasoning, PAL and
Program-of-Thoughts decouple computation from reasoning by emitting Python
code and delegating arithmetic to the interpreter,  which sidesteps the
well-known unreliability of LLM arithmetic \citep{gao2023pal,chen2023pot}.
Knowledge-graph (KG) executors play an analogous role for factual reasoning:
Think-on-Graph treats the LLM as an agent that interactively beam-searches
over KG triples, grounding each inferential hop in a retrieved KG edge
\citep{sun2024thinkongraph}, while KG-GPT segments claims, retrieves
candidate paths, and reasons over those structured paths
\citep{kim2023kggpt}. More broadly, tool-augmented language models such as Toolformer
internalise application programming interface (API) calls into the generation process \citep{schick2023toolformer},
and Self-RAG, a retrieval-augmented generation (RAG) method, integrates on-demand retrieval with explicit reflection tokens to
gate when and what evidence to incorporate \citep{asai2024selfrag}; the
research programme is surveyed in \citet{mialon2023augmented}.

In control-theoretic terms, external executors furnish deterministic evidence
about transitions that fall within their formalized operation or specification,
rather than a ground-truth oracle for the complete task. They remain
domain-specific and brittle to formalisation errors. What
remains uncontrolled is everything outside that domain: claims that cannot be
expressed in the symbolic formalism, retrievals that miss the relevant
evidence, and code paths whose tests are incomplete. The dominant failure
mode therefore shifts from miscalibrated self-judgment (\S\ref{sec:model-internal-validation})
to specification gaps between the natural-language task and the executable
artefact the model produces.

\subsubsection{Constraint-guided Generation}
\label{sec:constraint-guided-generation}

A complementary control component is to prevent invalid transitions from
being emitted at all, by intersecting the model's next-token distribution
with a constraint at decode time. Grammar-constrained decoding (GCD) realises
this for arbitrary context-free output spaces: \citet{geng2024grammar} show
that GCD substantially outperforms unconstrained decoding on structured natural language processing (NLP)
tasks without finetuning,  by masking tokens that cannot extend any sentential
form of the target grammar. \citet{willard2023efficient} formalise the
problem as transitions in a finite-state machine over the LM's vocabulary,
yielding the \texttt{Outlines} library  and an $O(1)$ amortised per-token
overhead,  while XGrammar accelerates this further by partitioning tokens into
context-independent and context-dependent classes and pre-computing masks,
reporting more than $10\times$ speedups over existing structured-generation
engines and near-zero overhead in some end-to-end settings for JSON and other
schemas~\citep{dong2024xgrammar}. PICARD pioneered the pattern in the
text-to-SQL setting, rejecting tokens that violate incremental SQL parsability
during beam search \citep{scholak2021picard}, and Synchromesh extends
constrained semantic decoding to richer formal languages, enforcing scope and
typing rules in addition to syntactic validity \citep{poesia2022synchromesh}.

Beyond pure grammar, several systems compile logical constraints directly
into the search. NeuroLogic decoding enforces predicate-logic constraints
over keywords and phrases during beam search \citep{lu2021neurologic}, and
NeuroLogic A$^{\ast}$esque augments this with lookahead heuristics that
estimate future constraint satisfaction, in the style of A$^{\ast}$
\citep{lu2022neurologic}. Programming-style abstractions raise the level of
expression: LMQL embeds constrained generation in a Python-like query
language with logical \texttt{where} clauses that the runtime compiles into
token-level masks and short-circuited inference, often reducing API cost in
addition to enforcing structure \citep{beurerkellner2023prompting}.
Abstractly, this yields a constrained token distribution of the form
\[
\tilde{p}(y_t \mid y_{<t}, x)
\propto
p_\theta(y_t \mid y_{<t}, x)\,\mathbf{1}[y_t \in \mathcal{A}(y_{<t})],
\]
where \(\mathcal{A}(y_{<t})\) denotes the set of admissible next tokens under
the active grammar or logical specification.

Within the trajectory-execution view, constraint-guided generation acts as a
\emph{decode-time admissible-action filter}: the controller can only choose
among transitions consistent with the specified output formalism, eliminating
malformed JSON or grammatically invalid SQL by construction; richer semantic
constraints can additionally block some typing or scope errors. What remains uncontrolled, however, is
\emph{semantic} correctness, a syntactically valid SQL query may still be
semantically wrong, and a schema-conforming JSON response may still be
factually false. There is also an expressivity hazard: \citet{banerjee2025crane}
show that restrictive final-answer grammars can reduce reasoning capability,
and propose augmenting the grammar so that intermediate reasoning remains
expressible while final answers conform to the target format.
This tension, strong constraints mask invalid actions but can also mask the
deliberation that produces correct ones, frames constraint-guided
generation as a component that controls form but not content, and that must
therefore be composed with the validation mechanisms of
\S\ref{sec:model-internal-validation} and \S\ref{sec:external-validation}.

\subsubsection{Multi-stage Validation Pipelines}
\label{sec:multistage-validation}

No validation source is reliable across all settings: model-internal
critique can be biased, executors are domain-bounded, and decode-time constraints
primarily control form. A growing body of work therefore treats validation as a
\emph{pipeline} that composes complementary signals along the trajectory.
The simplest such pipeline is generate--verify--refine: \citet{madaan2023self}
iterate generation, self-feedback, and revision with the same LLM, while
\citet{welleck2023selfcorrection} train a separate corrector that learns to
edit a frozen base generator using either scalar or natural-language
feedback.  CRITIC instantiates a tool-mediated variant in which an LLM
critiques its own output by issuing tool calls whose results then drive
revision, restoring the missing external signal that pure self-critique lacks
\citep{gou2024critic}. Neuro-symbolic systems such as Logic-LM and LINC are
themselves pipelines, alternating between LLM formalisation, symbolic
execution, and refinement loops driven by solver error messages
\citep{pan2023logiclm,olausson2023linc}.

A second pattern aggregates over multiple agents or samples. Multi-agent
debate has multiple LLM instances generate, critique, and revise responses
across rounds, yielding measurable gains in factuality and reasoning
\citep{du2023improving} and, in divergent variants, surfacing genuinely
distinct hypotheses for adjudication \citep{liang2024encouraging}. Verifier
ensembles operate on the trajectory rather than the agent: \citet{cobbe2021training}
introduced learned outcome verifiers for math word problems, used to rerank
candidate solutions,  and process reward models (PRMs) generalise this to
step-level scores that can be aggregated along a trajectory
\citep{lightman2023let,wang2024mathshepherd}. These signals compose
naturally with sampling: \citet{snell2024scaling} show that compute-optimal
allocation between sampling and verifier-guided search can outperform larger
models in the tested settings, while \citet{brown2024large} show that repeated
sampling increases the coverage of correct solutions over large inference
budgets. Interpreting the downstream verifier as a potential bottleneck is our
survey-level synthesis rather than a result directly established by either study.

In the control-centric reading, pipelines combine complementary validators
to compensate for the limitations of any single one: a PRM may catch errors
a grammar mask cannot see, an interpreter may catch errors a PRM misses, and
debate may expose errors invisible to a single rollout. What
remains uncontrolled is \emph{failure correlation across stages}: if all
validators draw on the same pretraining distribution or share a blind spot
(e.g.\ a flawed unit test that mirrors a flawed generator assumption), errors
propagate through the pipeline rather than being caught by it. There is
also a verifier--generator alignment problem: imperfect reward signals can
be exploited by best-of-$N$ search~\citep{yu2025scaling}. There is also a compute--reliability frontier: marginal returns to additional
verification stages diminish, while latency, cost, and error compounding
grow with depth. Designing pipelines is therefore itself a control problem,
in which the choice of which validators to compose, in what order, and at
what budget shapes the closed-loop reliability of trajectory execution. This
bridge is especially important for the next subsection: once validation is
partial, delayed, or ambiguous, uncertainty management determines whether the
controller should commit, branch, defer, or escalate verification.

\noindent\textbf{Representative domains.}
Domains with hard correctness constraints make the validation component indispensable. In automated theorem proving, outputs are formal proof scripts that a proof assistant such as Lean, Coq, or Isabelle must accept, so the language model acts as a proposal generator inside a verifier-in-the-loop pipeline rather than a standalone solver~\citep{polu2020generative, xin2024deepseek}. Neuro-symbolic systems apply the same pattern to logical reasoning, checking intermediate implications against a symbolic theory or delegating them to a solver~\citep{tafjord2020proofwriter, pan2023logiclm}. In cyber-physical requirements engineering, formalized specifications can be analyzed with SMT and model-checking tools; this illustrates the external validation that an LLM-based pipeline would require, although the cited study is not itself an LLM system~\citep{mavridou2020ten}. Across these domains, executable validators complement fluent derivations by rejecting candidates that violate the formalized specification.

\subsection{Uncertainty Management} \label{sub:uncertainty}

Uncertainty management dominates when a problem is ambiguous, underspecified, out of distribution, or when the available verifier is itself unreliable. The bottleneck there is calibrated commitment, and forcing the model to produce a single confident answer amplifies overconfidence; the appropriate control is to estimate uncertainty and to spend search or verification budget, or to abstain, in proportion to it~\citep{liu2024uncertainty, kadavath2022language}.

Even when explicit state, structured transitions, search, and validation are available, complex problem solving remains fundamentally uncertain: the controller may not know whether a candidate trajectory is correct, whether a validator is trustworthy in the current regime, or whether additional compute would meaningfully reduce error. The statistical structure of that uncertainty informs which action is useful together with the loss and available actions. Writing $H$ for a latent hypothesis or model state in an analytical predictive model, the law of total variance splits the predictive uncertainty of an answer $y$ into two parts,
\begin{equation}
\underbrace{\mathrm{Var}(y\mid x)}_{\text{total}}
= \underbrace{\mathbb{E}_{H}\!\left[\mathrm{Var}(y\mid x,H)\right]}_{\text{aleatoric (irreducible)}}
+ \underbrace{\mathrm{Var}_{H}\!\left(\mathbb{E}[y\mid x,H]\right)}_{\text{epistemic (reducible)}} .
\label{eq:total-variance}
\end{equation}
Under the standard interpretation in which $H$ indexes reducible model or hypothesis uncertainty, the first term represents residual variation under $H$ and the second variation across hypotheses. Retrieval or new evidence may reduce some epistemic uncertainty; additional sampling alone reduces Monte Carlo estimation error but does not supply missing knowledge. This decomposition is distinct from the sampling variance in \eqref{eq:bias-var-noise}: repeated rollouts can reduce estimator variance without necessarily reducing epistemic uncertainty about the task. A problem--control fit on this component therefore means directing information gathering and verification toward uncertainty that is plausibly reducible, while seeking clarification or abstaining when ambiguity remains irreducible. From this perspective, uncertainty management is not merely a matter of attaching confidence scores to final answers. It is a way of regulating commitment under ambiguity: uncertainty signals inform whether the system should commit, branch, invoke stronger validation, defer, or abstain, but the optimal action depends on the full belief and loss, not on a single spread statistic alone. The central question is thus not only how to estimate the two components of \eqref{eq:total-variance}, but how to turn that estimate into a policy over search, verification, and stopping \citep{shorinwa2024survey,liu2024uncertainty}.

\subsubsection{Diversity-based Uncertainty Estimation}

Diversity-based approaches estimate uncertainty by measuring how unstable the model's solution is across multiple sampled trajectories. Rather than relying on a single rollout, the controller samples several candidate reasoning paths and treats agreement among their induced answers as a proxy for confidence. Self-consistency is the canonical example: instead of committing to one chain-of-thought, the model samples diverse trajectories and aggregates their answers, improving performance on arithmetic and commonsense reasoning tasks by exploiting redundancy across valid paths \citep{wang2022self}. In control terms, the key benefit is not only better final accuracy, but an observable estimate of how fragile the current decision is under stochastic variation in the reasoning process.

Formally, let \(y^{(1)}, \dots, y^{(N)}\) denote the answers induced by \(N\) sampled trajectories for input \(x\). The empirical answer distribution is
\[
\hat{p}(y \mid x) = \frac{1}{N}\sum_{i=1}^{N}\mathbf{1}[y^{(i)} = y],
\]
and a simple disagreement-based uncertainty score is
\[
u(x) = 1 - \max_y \hat{p}(y \mid x).
\]
Low \(u(x)\) indicates that one answer dominates across rollouts, while high \(u(x)\) indicates that the reasoning process is unstable and may warrant further verification, branching, or delayed commitment. More broadly, inference-time scaling results suggest that additional sampling can systematically expand coverage of correct solutions, making agreement rates and answer dispersion increasingly informative control signals when the model has access to enough compute \citep{brown2024large,snell2024scaling}.

From a control perspective, diversity-based uncertainty estimation is attractive because it is label-free and naturally integrated with sampling-based search from Section~\ref{sub:search}. The same rollouts that improve answer quality can also expose when the controller is operating in an unstable region of the problem space. What remains uncontrolled, however, is correlated error across samples: multiple trajectories may still share the same latent misconception, so high agreement does not guarantee correctness. Moreover, disagreement localizes uncertainty only at the level of whole trajectories or final answers; it does not by itself identify which intermediate decision caused the instability. Thus, diversity can reveal that commitment is risky, but not necessarily how the trajectory should be repaired.

\subsubsection{Confidence-aware Decision Making}

Another line of work estimates uncertainty through explicit confidence signals and uses those signals to gate downstream decisions. In the simplest setting, the controller associates each input with a confidence score \(c(x)\) and chooses whether to answer directly, request additional verification, or abstain. This converts uncertainty from a descriptive statistic into an action-selection criterion:
\[
\hat{y} =
\begin{cases}
\arg\max_y p_\theta(y \mid x), & c(x) \ge \tau_c, \\
\text{abstain / verify / defer}, & c(x) < \tau_c,
\end{cases}
\]
where \(\tau_c\) is a decision threshold. The central challenge is that raw model confidence is often poorly calibrated, so the usefulness of this rule depends on whether \(c(x)\) tracks actual correctness rather than merely the sharpness of the next-token distribution \citep{jiang2021can,guo2017calibration,nixon2019measuring,kadavath2022language}.

Recent LLM-specific work has explored stronger confidence surrogates. DeepThink-with-Confidence uses model-internal confidence signals to filter low-quality reasoning traces during or after generation, improving the allocation of inference-time compute across candidate paths \citep{fu2025deepthinkconfidence}. Supervised uncertainty estimators for LLMs similarly attempt to learn confidence signals that better correlate with downstream correctness than naive token probabilities alone \citep{liu2024uncertainty}. These methods are especially useful when the system must make selective decisions, such as whether to present an answer to a user, whether to trigger a verifier, or whether to route the instance to a more expensive reasoning pipeline.

From a control perspective, confidence-aware decision making provides a direct mechanism for selective commitment: the controller can refuse to act when its own estimate of reliability is too low. This matters in high-stakes settings where a wrong but confident answer is more harmful than a deferred one. What remains uncontrolled is calibration drift. Confidence estimates that appear well behaved on benchmark distributions may become unreliable under task shift, prompt variation, or adversarial perturbation, causing the controller to commit too early or defer unnecessarily. Confidence is therefore useful only to the extent that it remains coupled to empirical correctness rather than to superficial fluency.

\subsubsection{Uncertainty-guided Control Policies}

Beyond deciding whether to answer, uncertainty can be used to modulate how much control effort the system allocates to an instance or an intermediate state. In this setting, uncertainty becomes a policy signal that determines search depth, verifier budget, branching width, or information-seeking behavior. A simple abstraction is
\[
B(x) = B_{\min} + \alpha\, u(x),
\]
where \(u(x)\) is an uncertainty estimate and \(B(x)\) is the compute budget assigned to input \(x\), for example the number of additional samples, the depth of search, or the number of verification calls. The controller therefore spends more effort precisely where the current trajectory is least trustworthy.

This idea already appears implicitly in inference-time scaling results, where additional compute is allocated through repeated sampling or verifier-guided search, and performance depends strongly on how that compute is distributed across candidate trajectories \citep{snell2024scaling,brown2024large}. More explicit uncertainty-aware search methods refine this by making branch expansion depend on value uncertainty rather than only on point estimates. For example, uncertainty-aware value models represent branch quality as a distribution and use that uncertainty during search, mitigating the scaling failures of verifier-guided search that arise when a noisy evaluator is treated as ground truth \citep{yu2025scaling,yu2025uncertainty}. At the agent level, uncertainty propagation work makes a related point: local uncertainty in intermediate decisions can accumulate across tool use and multi-step planning, so control policies must account for compounding uncertainty rather than only one-step confidence \citep{zhaoetal2025uncertainty}.

From a control perspective, uncertainty-guided policies are valuable because they close the loop between estimation and action: uncertainty no longer merely diagnoses a problem after the fact, but actively changes the controller's behavior. This is particularly important in long-horizon reasoning, where uniform compute allocation is wasteful and premature commitment to low-confidence states can destabilize the entire trajectory. What remains uncontrolled, however, is the reliability of the uncertainty estimate itself. If \(u(x)\) is delayed, biased, or miscalibrated, the controller may spend its budget in the wrong places, widening search where none is needed or committing early when stronger validation was required.

\subsubsection{Probabilistic Control Signals}

The most explicit formulation treats uncertainty as a first-class probabilistic object within the control loop rather than as an auxiliary heuristic. Instead of attaching a single confidence score to the final answer, the controller reasons over predictive distributions, token-level uncertainty, or value distributions associated with actions and trajectories. In that setting, action selection can be written as a risk-sensitive objective,
\[
a_t^\ast
=
\arg\max_{a \in \mathcal{A}(s_t)}
\mathbb{E}[R(s_t,a)] - \lambda\,\mathrm{Risk}(s_t,a),
\]
where \(R(s_t,a)\) denotes expected utility and \(\mathrm{Risk}(s_t,a)\) penalizes uncertain or high-variance transitions. The trade-off parameter \(\lambda\) controls how aggressively the system sacrifices expected reward to avoid brittle or poorly understood trajectories.

Several recent directions instantiate parts of this view. Token-level uncertainty estimation aims to identify which segments of a reasoning trace are most unreliable, enabling fine-grained control over where validation or revision should be applied \citep{zhang2025token}. Uncertainty-aware value models similarly replace point estimates over branch quality with predictive distributions, allowing the search policy to distinguish promising-but-uncertain branches from confidently poor ones \citep{yu2025uncertainty}. More broadly, supervised uncertainty quantification methods seek to produce probabilistic confidence estimates that remain informative for downstream control, rather than only for post hoc evaluation \citep{liu2024uncertainty,shorinwa2024survey}.

From the estimation-and-decision view of this survey, probabilistic control signals are the most explicit formulation of uncertainty management because they can, in principle, unify commitment, search, validation, and stopping under a common decision rule. Their main limitation is practical rather than conceptual: accurate probability estimates are difficult to obtain for long reasoning trajectories, and errors in uncertainty estimation can themselves propagate through the control loop. A probabilistic controller is therefore only as reliable as the distributions it acts on. When those distributions are poorly calibrated, incomplete, or fragile under distribution shift, the resulting policy may appear principled while still making systematically bad commitments. This leaves calibration, distributional robustness, and uncertainty propagation as central unresolved problems for uncertainty-aware CPS.

\noindent\textbf{Representative domains.}
Uncertainty management dominates in domains where inputs are ambiguous or answers are not uniquely determined. CommonsenseQA and SocialIQA contain commonsense and social-inference questions with competing plausible interpretations, making calibrated selection relevant even though the benchmark annotations specify a target answer~\citep{talmor2018commonsenseqa,sap2019socialiqa}. Mathematical reasoning under distribution shift is a second case: perturbation benchmarks such as GSM-PLUS report large accuracy drops under meaning-preserving edits, and TreeCut studies how models should respond to underspecified or unanswerable mathematical problems~\citep{Li2024GSMPLUS,Ouyang2025TreeCut}. These settings motivate calibrated confidence and selective abstention; aggregation helps only when the sampled answer distribution carries useful information about correctness.

\section{Cross-Cutting Dimensions: Distributing and Internalizing Control}
\label{sec:crosscutting}

The five components of Section~\ref{sec:taxonomy} describe a single estimation-and-decision loop. Two further dimensions cut across that loop and compose with any of the components rather than adding a new one. Control may be distributed across several agents, so that state representation, validation, and the remaining components are divided among specialized roles, and control may be internalized into model weights at training time rather than supplied by external scaffolding. In the statistical reading of this survey these are two ways of realizing the same filter: distributing control approximates the belief by an ensemble of local estimators whose messages play the role of a consensus update, and may amortize selected parts of the proposal, evaluation, or decision policy. We treat these dimensions separately because they reshape how the core components are realized rather than introducing a new kind of control.

\subsection{Distributed and Multi-Agent Control}\label{sub:distributed}

Distributed control addresses CPS settings in which a single controller is inadequate because relevant state, constraints, or failure signals are naturally partitioned across subtasks, tools, memories, or agents. In this setting, control is realized by multiple local controllers that each operate on partial state and exchange messages to maintain consistency of the global trajectory. A generic formulation is
\[
x_i^{(t+1)} = f_i\!\left(x_i^{(t)},\, u_i^{(t)},\, m_i^{(t)}\right),
\qquad
u_i^{(t)} = \pi_i\!\left(x_i^{(t)},\, m_i^{(t)}\right),
\]
where \(x_i^{(t)}\) is the local state of controller \(i\), \(u_i^{(t)}\) is its local control action, and
\[
m_i^{(t)} = \mathcal{M}_i\!\left(\{x_j^{(t)},u_j^{(t)}\}_{j\in \mathcal{N}(i)}\right)
\]
aggregates information received from neighboring controllers. The global objective is not merely to generate multiple reasoning traces, but to regulate interactions so that local decisions remain compatible with shared constraints and the overall process remains recoverable under partial observability.

From the control perspective of this survey, distributed control does not add a new loop component but a cross-cutting dimension, the \emph{decomposition of control authority}. Instead of relying on a single monolithic reasoning loop, the system allocates proposal, checking, planning, retrieval, or execution to separate modules and then coordinates them through explicit communication. This can reduce local overload and expose disagreements that would remain hidden in single-agent generation. However, decomposition alone does not guarantee stable global behavior. The main failure mode is that local controllers may be individually coherent while collectively inconsistent, causing conflict, duplication, or premature convergence to a wrong trajectory. The key question is therefore not whether multiple agents are present, but what aspects of the trajectory they actually control and what remains uncontrolled.

In this survey, we mainly use distributed control in a broad functional sense, including decentralized controllers and centrally orchestrated multi-agent systems that distribute control authority across modules. The following four patterns can be understood as non-exclusive dimensions of distributed control; a single architecture can combine role specialization, interaction-based correction, communication protocols, and hierarchical coordination.

\subsubsection{Role-specialized Decomposition}

A first pattern decomposes control by assigning different subtasks to different controllers. Systems such as CAMEL, AutoGen, MetaGPT, and ChatDev separate roles such as planner, coder, reviewer, or user proxy, thereby turning a single reasoning loop into a distributed pipeline of local policies \citep{li2023camel,wu2024autogen,hong2024metagpt,qian2023chatdev}. The control component here is \emph{functional specialization}: each controller governs a restricted part of the state update, for example task decomposition, code generation, or critique. This addresses the failure of single-controller overload, where one model must simultaneously maintain task state, generate candidate actions, and evaluate their validity.

Formally, if the global state is partitioned as \(x^{(t)} = (x_1^{(t)},\dots,x_K^{(t)})\), then specialized control can be written as
\[
u_i^{(t)} = \pi_i\!\left(x_i^{(t)}, m_i^{(t)}\right),
\qquad
x^{(t+1)} = F\!\left(x^{(t)}, u_1^{(t)},\dots,u_K^{(t)}\right).
\]
This decomposition improves observability of intermediate decisions and makes failure localization easier, since errors can often be traced to a particular role. What remains uncontrolled is the consistency of the composition \(F\). Role specialization does not by itself ensure that local outputs are compatible, that interfaces are well aligned, or that an early mistake passed from one role to another will be corrected rather than amplified.

\subsubsection{Interaction-based Correction}

A second pattern uses interaction itself as a control mechanism. Debate, critique, and reconciliation protocols let one controller challenge or revise another controller's proposal, creating a distributed correction loop rather than a single-pass transition \citep{du2023improving,chen2023reconcile,irving2018ai,brown2023scalable}. The control component is therefore \emph{cross-checking by disagreement}: instead of committing immediately to one transition, the system allows competing local updates and uses interaction to expose contradictions or missing evidence. This addresses the failure mode of unchallenged local error, in which a plausible but incorrect proposal persists simply because no independent controller inspects it.

A simple abstraction is
\[
\tilde{u}_i^{(t)} = \pi_i\!\left(x_i^{(t)}, m_i^{(t)}\right),
\qquad
u^{(t)} = \Gamma\!\left(\tilde{u}_1^{(t)},\dots,\tilde{u}_K^{(t)}\right),
\]
where \(\Gamma\) is a reconciliation operator such as voting, judging, or critique-based revision. Interaction improves local error detection, but what remains uncontrolled is the reliability of \(\Gamma\). If the aggregation rule rewards fluency rather than correctness, or if all controllers share similar biases, then distributed interaction can still converge to a confidently wrong answer. In that case, the system has more discussion, but not more effective control.

\subsubsection{Communication and Coordination Protocols}

A third pattern focuses on how local controllers exchange state. Long-context and communication-oriented systems such as Chain-of-Agents and LongAgent distribute reading, summarization, and reasoning across multiple workers so that each controller handles only part of the available information \citep{zhang2024chain,zhao2024longagent}. The control component is \emph{message passing over partial observations}. This addresses the failure of limited context and local visibility: no single controller has to process the entire problem state at once, and intermediate summaries can be propagated to support downstream decisions.

The underlying control problem can be written as a consensus-like objective:
\[
\min_{\{u_i\}} \sum_{i=1}^K \ell_i(x_i,u_i)
\quad \text{subject to} \quad
C\!\left(x_1,\dots,x_K\right)=0,
\]
where \(C(\cdot)=0\) denotes consistency constraints induced by communication. In practice, these methods improve scalability and partial-state coverage, but what remains uncontrolled is information loss through compression and communication bias. Messages may omit exactly the evidence needed to overturn an incorrect trajectory, and local controllers may overweight shared summaries while underusing unique local information. This is especially problematic in hidden-profile settings, where correct control requires pooling distributed evidence rather than repeatedly reinforcing what is already shared \citep{li2025assessing}.

\subsubsection{Hierarchical Distributed Control}

A fourth pattern introduces hierarchy, in which a higher-level controller allocates subgoals or delegates work to lower-level controllers, while lower-level controllers execute local transitions under that allocation. This pattern appears in modular agent systems and in socially grounded agent simulations where planning and execution are separated across levels \citep{park2023generative,zhou2023sotopia,qian2024scaling}. The control component is \emph{multi-level regulation}: high-level controllers determine decomposition and resource allocation, while low-level controllers handle local execution. This addresses the failure of flat coordination, where too many peer-to-peer interactions can create drift, redundancy, or unstable switching among candidate trajectories.

A hierarchical decomposition can be written as
\[
g^{(t)} = \Pi_{\mathrm{hi}}\!\left(x^{(t)}\right),
\qquad
u_i^{(t)} = \pi_{i,\mathrm{lo}}\!\left(x_i^{(t)}, g^{(t)}, m_i^{(t)}\right),
\]
where \(g^{(t)}\) is a high-level subgoal or routing decision. This improves control over long-horizon structure by separating global planning from local execution. What remains uncontrolled, however, is whether the hierarchy itself is adaptive and whether lower-level failures are propagated upward in time to trigger replanning. If the high-level controller commits to a poor decomposition, the hierarchy can efficiently execute the wrong plan.

Overall, distributed control expands the control surface of CPS by decomposing authority across multiple local controllers, communication channels, and levels of abstraction. Its main benefit is not collective reasoning in the abstract, but explicit regulation of who controls which part of the trajectory. Its main limitation is that decomposition does not by itself supply calibrated aggregation, faithful communication, or principled recovery from collective error. Thus, distributed control should be viewed as a way to allocate and expose control, not as a substitute for validation, uncertainty management, or rollback.

\subsection{Training-Time Internalization of Control}\label{sub:training}

This section examines how training may support the partial internalization of control. In practice, two strategies are especially important: fine-tuning, particularly on curated datasets designed to elicit reasoning \citep{wu2025llm}, and reinforcement learning, which shapes model behavior through reward signals \citep{cao2024survey}. From the perspective of CPS, these post-training methods matter not only because they can improve reasoning performance, but also because they may reduce the need for some forms of external control at inference time.

\subsubsection{Supervised Internalization of Reasoning Processes}

Supervised fine-tuning amortizes the base proposal component of the transition mechanism: it learns a prior over reasoning traces by training the model on trajectories, intermediate steps, or task-specific decompositions \citep{wu2025llm, zelikman2022star}, so that part of the proposal behavior that external scaffolding would otherwise shape is folded into the weights. When supervision includes step-by-step rationales, the learned prior favors more explicit intermediate states and locally coherent transitions rather than unconstrained next-token continuation. In estimation terms, this reduces proposal error relative to the supervised task distribution, but it does not by itself supply a likelihood or a decision rule, so validation and calibrated commitment remain external.

This helps address a common failure mode of base models: they may generate fluent responses without maintaining a stable problem representation or a controlled reasoning trajectory. By learning from curated reasoning data, the model becomes more likely to follow intermediate structures that resemble valid solution progress. However, this internalization remains partial. Supervised traces can teach the model to imitate controlled reasoning patterns, but they do not guarantee faithfulness, robust recovery from early mistakes, or reliable behavior under distribution shift. As a result, supervised internalization can strengthen local control, while leaving global verification and correction only weakly internalized.

 Representative examples appear at multiple levels. Rationale-rich supervision makes intermediate solution structure more explicit, while domain-specific fine-tuning biases the model toward task-relevant decompositions. Continual learning extends this idea by preserving such structured behaviors across tasks \citep{zheng2025lifelong}, and systems such as \textbf{Voyager} suggest how reusable skills can be carried into more open-ended environments \citep{wang2023voyager}.
These cases show that supervision can shape not only answer quality but also the default reasoning patterns the model is likely to follow. At the same time, the learned control remains limited by the scope and quality of the supervision, and may still break under distribution shift or stronger recovery demands.

\subsubsection{Preference- and Reward-based Policy Shaping}

Preference- and reward-based training internalizes a decision rule: it learns a utility surrogate over trajectories and biases the policy toward the outputs that maximize it \citep{cao2024survey, bai2022training, chaudhari2024rlhf}. Where supervised fine-tuning amortizes the proposal, these methods amortize part of the decision policy induced by $L$ \eqref{eq:bayes-decision} into the weights, so that the choice of which trajectory to favor is made by the policy rather than by an external decision layer. When the reward is a step-level process signal rather than an outcome, the signal can be interpreted as partially shaping a likelihood surrogate when it tracks transition validity and is incorporated into the policy.

A standard formulation is reinforcement learning from human feedback (RLHF), where a reward model $R_\phi(x,y)$ scores the response $y$ to input $x$, and the policy $\pi_\theta$ is optimized to maximize reward while staying close to a reference policy $\pi_{\mathrm{ref}}$ through Kullback--Leibler (KL) regularization:
\begin{equation}
\max_{\theta}\;\; \mathbb{E}_{x\sim\mathcal{D},\,y\sim \pi_\theta(\cdot\mid x)}\Big[R_\phi(x,y)\Big]
\;-\;\beta\,\mathrm{KL}\!\left(\pi_\theta(\cdot\!\mid\! x)\,\|\,\pi_{\mathrm{ref}}(\cdot\!\mid\! x)\right).
\label{eq:rlhf-kl}
\end{equation}
This helps address an important failure mode of purely supervised internalization. A model may be able to produce plausible intermediate steps, yet still prefer trajectories that are unhelpful, weakly justified, unsafe, or misaligned with task objectives. Preference-based optimization can push the policy toward responses that better satisfy correctness, coherence, helpfulness, or constraint-related requirements. In reasoning tasks, this includes favoring trajectories that are more verifiable, better formatted for downstream checking, or more consistent with human judgments of valid reasoning \citep{ouyang2022training}.
Representative examples illustrate different reward sources. In RLHF, the reward is learned from human preferences \citep{bai2022training, ouyang2022training}. In other settings, rewards can be programmatically defined, such as rewarding correct math answers or code that passes execution tests. Recent reasoning-oriented systems such as DeepSeek-R1 suggest that such automated rewards can encourage the emergence of more explicit multi-step reasoning when these behaviors help maximize reward \citep{guo2025deepseek}. Process supervision extends this idea further by rewarding not only final outcomes but also properties of the intermediate reasoning trajectory \citep{ye2025beyond}.

However, the resulting control remains partial. If rewards are incomplete, outcome-only, or poorly aligned with the intended reasoning process, the model may exploit shortcuts or optimize evaluator artifacts rather than maintain faithful control over the trajectory. Reward shaping can therefore internalize preference over trajectories, but it does not by itself guarantee robust verification, reliable recovery from early mistakes, or faithful reasoning under distribution shift.

\subsubsection{Optimization Objectives and Stability Constraints}

Optimization objectives are regularized risk minimization on the policy: they maximize the learned utility while constraining how far each update moves the policy, which is what keeps the amortization of the decision rule stable. Once reward- or preference-based signals are introduced, the remaining question is how to update the policy without unstable drift, reward overfitting, or collapse toward narrow behaviors, so the objective couples a risk term to an explicit or implicit trust region. In this sense optimization objectives and stability constraints govern the internalization process itself.

A representative example is Proximal Policy Optimization (PPO), which updates the policy to increase reward while constraining how far each update moves from the policy that generated the data, using a clipped likelihood ratio and a KL penalty to a reference policy~\citep{schulman2017proximal}. Under the CPS lens this is update-level control: the objective improves reward while limiting how aggressively new behavior is absorbed into the policy.

\noindent\textbf{Direct Preference Optimization (DPO).} DPO removes the rollout and critic overhead of PPO by training directly on preference pairs rather than performing online reinforcement learning~\citep{rafailov2023direct}. It raises the model's relative likelihood of preferred over dispreferred responses while staying close to a reference policy, providing a simpler offline optimization route to bias the policy toward preferred reasoning trajectories.

\noindent\textbf{Group Relative Policy Optimization (GRPO).} GRPO keeps exploratory reinforcement learning but eliminates the learned critic, replacing it with a group baseline: for each prompt it samples several responses and uses each reward minus the group mean as a relative advantage~\citep{shao2024deepseekmath}. This lowers memory and compute and suits math and code reasoning, where programmatic rewards are available but training a value function is costly.

Some other methods, such as Odds Ratio Preference Optimization (ORPO), Kahneman--Tversky Optimization (KTO), and lightweight offline preference losses such as Rank Responses to Align Human Feedback (RRHF) and Sequence Likelihood Calibration with Human Feedback (SLiC-HF), push further on simplicity and data efficiency. ORPO removes the explicit reference policy by replacing KL regularization with an odds-ratio penalty, which reduces dependency on a strong baseline but can require careful tuning for stability \citep{hong2024orpo}. KTO learns from binary desirable--undesirable feedback through a prospect-theoretic objective. Its standard formulation retains a reference policy and uses an estimate of the KL divergence as a reference point, enabling training from unpaired and highly imbalanced feedback \citep{ethayarajh2024model}. RRHF and SLiC-HF adopt simple ranking or margin losses over preferred versus dispreferred responses with optional length or verbosity controls; they are easy to scale offline but, like DPO, inherit the limits of the preference corpus and do not provide on-policy exploration \citep{yuan2023rrhf,zhao2023slic}.

\subsubsection{Alignment as Learned Constraint Enforcement}

Alignment installs a learned constraint prior rather than a guaranteed posterior validation: training makes invalid or undesired trajectories less likely by default, but it does not check any particular trajectory against evidence at inference time. Under the CPS lens, fine-tuning and reinforcement learning encourage the model to default to behaviors that are more helpful, truthful, harmless, and consistent with user intent \citep{ouyang2022training, guo2025deepseek}, which lowers the prior mass on undesired trajectories. This is why alignment reduces but cannot eliminate the need for external validation: a prior that downweights bad trajectories is not the same as an observation model that rejects a specific bad one.

This helps address an important limitation of purely performance-oriented reasoning optimization. A model may generate accurate answers yet still rely on unfaithful explanations, unsafe reasoning patterns, or behaviors misaligned with user expectations. Alignment-oriented supervision and reward design push the policy toward trajectories that are not only effective but also more trustworthy and norm-consistent. However, this internalization remains incomplete. Constraints learned during training may still fail under novel tasks, long-horizon reasoning, or adversarial conditions. Alignment should therefore be understood not as a replacement for external control, but as a partial way of embedding constraint-following behavior into the model.

\section{Benchmarks and Evaluation}
\label{sec:benchmarks}

Read statistically, evaluation is the problem of estimating which quantity a metric actually measures. The empirical error rate estimates the 0--1 risk
\[
R=\mathbb{E}[\mathbf{1}\{\hat y\neq y^\star\}],
\]
whereas accuracy estimates \(1-R\). Either outcome statistic collapses the error structure summarized by \eqref{eq:bias-var-noise} into a single aggregate and therefore cannot identify whether an observed failure is dominated by systematic bias, sampling variability, irreducible ambiguity, or divergence of the intermediate state estimate. Because our concern is whether the solution process is controlled, evaluation should also probe the components of the loop directly: whether the state is a sufficient statistic, whether the proposal kernel is well formed, whether the observation model reweights the belief correctly, whether the system recovers from error, and whether the reported uncertainty is calibrated. This section contrasts outcome-level and process-level evaluation, reviews common benchmarks, maps each metric to the statistical quantity it estimates (Table~\ref{tab:tab5}), and discusses what current benchmarks fail to measure. More strongly, evaluation should not only rank methods but diagnose problem--control fit: a benchmark that reports only a risk point estimate cannot indicate which component of the decomposition an intervention should target.

\subsection{Outcome vs. Process Evaluation}

Most widely used benchmarks score only the final answer. Outcome-level evaluation is attractive because it is simple, objective, and comparable across models, but it is a weak proxy for whether the underlying process was controlled. A model can reach the correct answer through memorization, spurious correlation, or a flawed chain whose errors happen to cancel, and outcome accuracy cannot separate these cases from sound reasoning~\citep{niu2024largelanguagemodelscognitive}. The gap widens on complex problems, where long trajectories, distribution shift, and the difficulty of generating genuinely novel instances make final-answer accuracy an unreliable signal of reasoning quality.

Process-level evaluation instead asks whether the trajectory itself is sound: whether intermediate steps are valid, whether they are faithful to the final answer, and whether errors are detected and corrected. Process signals can also support control at inference time. DeepConf illustrates this use by employing model-internal confidence signals to filter low-quality reasoning trajectories during or after generation, with reported improvements in accuracy and computational efficiency~\citep{fu2025deepthinkconfidence}. It should therefore be understood as an inference-time control method that uses process information, rather than as a benchmark or metric for process-level evaluation. More generally, the availability of a process signal does not by itself establish that the exposed rationale faithfully represents the model's latent computation; evaluating the process still requires explicit criteria for transition validity, faithfulness, error detection, and recovery. The remainder of this section organizes evaluation around the statistical quantity each metric estimates, so that a benchmark or metric is judged by which part of the loop, and which term of the decomposition, it actually measures.

\subsection{Common Benchmarks and Evaluation Suites}

Benchmarks for complex problem solving are often described as domain-specific, broad-coverage, or reasoning-focused, although these categories overlap. GSM8K and MATH, for example, are mathematics benchmarks but also test multi-step reasoning. The distinction is therefore useful for describing what a benchmark emphasizes, not for assigning it to a single class.

Domain-specific benchmarks concentrate on fields such as mathematics, medicine, or law. GSM8K contains 8.5K grade-school mathematics word problems~\citep{cobbe2021training}, whereas MATH consists of more difficult competition-style problems~\citep{hendrycks2021measuring}. Both require domain knowledge and multi-step solution construction. Their standard scoring, however, is based mainly on final answers and reveals little about the validity of intermediate states or transitions.

Broad-coverage benchmarks examine performance across subjects and task formats. Massive Multitask Language Understanding (MMLU) spans 57 subjects, including mathematics, history, computer science, and law~\citep{hendrycks2020measuring}, and MMLU-Pro provides a more difficult, reasoning-intensive extension~\citep{wang2024mmluprorobustchallengingmultitask}. BIG-Bench Hard collects 23 tasks that were challenging for earlier language models~\citep{suzgun2022challengingbigbenchtaskschainofthought}. Such benchmarks expose variation across tasks, but performance across many benchmark domains does not by itself demonstrate generalization to unseen domains or robustness under distribution shift.

Reasoning-focused benchmarks place greater emphasis on particular forms of inference, including logical, causal, analogical, commonsense, and mathematical reasoning. MME-Reasoning evaluates inductive, deductive, and abductive reasoning in multimodal models while controlling for selected perceptual and knowledge confounds~\citep{yuan2025mmereasoningcomprehensivebenchmarklogical}. Even here, final-answer scoring provides only indirect evidence about the reasoning process. It does not show whether the system maintained an adequate state, validated its transitions, detected errors, or recovered from them.

Together, these benchmarks provide useful task coverage, but they do not directly measure control. In the present framework, benchmarks specify the problems on which a system is tested; the metrics and experimental protocol determine which aspects of the control loop are observed. Process annotations, controlled perturbations, recovery tests, and matched inference budgets are therefore needed to complement conventional outcome-level evaluation. The next subsection organizes these metrics by the statistical quantities they estimate and the components of the control loop they probe. Table~\ref{tab:tab5} uses expected calibration error (ECE) as a calibration metric.

\subsection{Metrics as Estimators}

\begin{table}[t]
\centering
\small
\caption{Mapping common evaluation signals onto the loop component they probe and the statistical quantity each estimates.}
\label{tab:tab5}
\begin{tabularx}{\textwidth}{@{}p{2.6cm}p{4.6cm}X@{}}
\toprule
Loop component & Representative evaluation signal & Statistical quantity estimated \\
\midrule
Explicit state         & trace inspectability; state-recovery probes & sufficiency of the state statistic \\
Transition structuring & step-level validity; decomposition correctness & bias of the proposal kernel \\
Validation             & execution accuracy~\citep{chen2021evaluating}; prover pass rate~\citep{lightman2023let} & risk under an executable loss; likelihood fit \\
Search / rollback      & self-consistency~\citep{wang2022self}; recovery-after-error rate & variance / dispersion of the answer estimator \\
Uncertainty            & calibration, ECE and Brier~\citep{kadavath2022language}; selective accuracy & proper scoring; risk--coverage of the retained set \\
Cross-cutting          & faithfulness~\citep{lanham2023measuring}; efficiency (tokens, latency) & causal sufficiency of the trace; cost per unit risk reduced \\
\bottomrule
\end{tabularx}
\end{table}

\noindent\textbf{Outcome-risk estimators.} The most common signals estimate the risk of the final answer, the expected loss $\mathbb{E}[\mathbf{1}\{\hat y \neq y^\star\}]$ and its relatives. Task accuracy, precision, recall, and the F1 score are point estimates of this risk under a $0$--$1$ or class-weighted loss, and they dominate leaderboard reporting for their simplicity and comparability \citep{bernado2003accuracy, goutte2005probabilistic}. Exact match against a reference is the same estimator restricted to a discrete answer space, used by GSM8K~\citep{cobbe2021training}, MATH~\citep{hendrycks2021measuring}, and SVAMP~\citep{pitis2023boosted}, and extended to commonsense, multi-hop, and reading comprehension by ARC-Challenge~\citep{clark2018think}, BIG-Bench Hard~\citep{srivastava2023beyond}, and DROP~\citep{dua2019drop}; it penalizes semantically correct answers in a different surface form. Execution accuracy replaces string comparison with a run of the predicted program or expression, estimating risk under an executable loss and so capturing functional rather than surface correctness \citep{chen2021evaluating}. What outcome-risk estimators cannot do is decompose that risk: a correct answer reached through a flawed trajectory scores identically to a sound one, so they cannot separate bias, variance, or noise \citep{wei2022chain, lyu2023faithful, ni2023lever}.

\noindent\textbf{State and process estimators.} A second group estimates properties of the trajectory rather than the answer, targeting the sufficiency of the state and the validity of transitions. Process-oriented evaluation asks whether intermediate steps are valid, faithful, and consistent with the final prediction, since a correct answer can be reached by a spurious shortcut \citep{bang2023multitask, zhang2023m3exam}. Step-level correctness estimates the validity of each transition against gold steps, formal checkers, or trained judges \citep{prasad2023receval, ma2023let}, gives localized error signals and partial credit, and underlies process supervision \citep{lightman2023let, luo2024improve}. Faithfulness estimators target causal sufficiency, whether the rationale is causally tied to the answer rather than a post-hoc narrative: perturbation tests delete or alter steps and check whether the answer moves \citep{lanham2023measuring}, grounding checks verify that each cited quantity is necessary and supported, and entailment- or prover-based faithfulness scores make the notion quantitative \citep{kirchner2024prover, huang2022large}. These estimators are limited by annotation cost, domain specificity, and the reliability of the automatic checker.

\noindent\textbf{Sampling-dispersion and stability diagnostics.}
Repeated stochastic sampling and answer-dispersion measures provide diagnostics of conditional output stability rather than direct estimates of the variance term in \eqref{eq:bias-var-noise}: by sampling independent chains and measuring their agreement, the consistency rate is a label-free estimate of how stable the answer is under sampling, with high dispersion signaling reliance on incidental heuristics \citep{wang2022self}. In program synthesis the same quantity is measured by whether candidate programs pass identical tests \citep{min2023beyond}. As Section~\ref{sub:search} makes explicit through the effective sample size of \eqref{eq:ess}, this estimator sees only variance: it cannot detect a bias shared across correlated samples, which is precisely the regime in which high consistency co-occurs with a confident wrong answer.

\noindent\textbf{Likelihood and verifier-quality estimators.} Process reward models and learned or symbolic verifiers estimate the fit of the observation model, how well an external signal scores intermediate transitions. A process reward model (PRM) assigns fine-grained scores to partial trajectories, so its average over a trace is a process metric and its gradient a training signal \citep{setlur2024rewarding}. Verifier-based evaluation uses natural language inference (NLI)-style checkers, symbolic solvers, unit tests, or prover--verifier setups to judge step validity and yield localized feedback \citep{lightman2023let, kirchner2024prover}. Because these estimators are themselves models, their value is bounded by their own calibration and coverage: a verifier with blind spots reports a misspecified likelihood, and gaps produce false positives and negatives.

\noindent\textbf{Decision-quality estimators.} A fifth group estimates whether the belief supports good decisions rather than whether the answer is correct. Calibration error, proper scoring rules such as the Brier score, risk--coverage curves, and selective accuracy measure the quality of the commit, verify, or abstain decision of \eqref{eq:bayes-decision} rather than accuracy alone \citep{kadavath2022language}. Current benchmarks report these rarely, which the next subsection treats as a systematic gap.

\noindent\textbf{Efficiency-adjusted estimators.} Finally, efficiency and robustness estimators normalize quality by cost and by input perturbation. Efficiency metrics, inference time, token count, path length, memory, and tool-call frequency, measure the resource footprint, and read against the risk reduced they estimate cost per unit risk reduced \citep{yao2023react, chang2024survey}. Robustness estimators measure the stability of the risk estimate under meaning-preserving perturbations such as synonym substitution, reordering, or distractor injection, where advanced models still degrade sharply \citep{liu2024exploring, jia2017adversarial, lanham2023measuring}. These estimators guard against reading a low risk on a narrow distribution as a low risk in general.

\subsection{Limitations of Current Benchmarks}

Beyond what they measure well, current benchmarks systematically under-measure several control-relevant properties: how well confidence is calibrated, whether reasoning transfers under distribution shift, and how systems perform when a human remains in the loop.

\noindent\textbf{Calibration.}
Calibration measures how well the probability estimates output by the model correspond to actual correctness. Ideally, a model predicting an answer with 80\% confidence should be correct about 80\% of the time. However, studies have shown that LLMs are systematically overconfident in reasoning tasks \citep{kadavath2022language, xu2025large, jiang2021can}. Poor calibration undermines trust in model predictions, especially in high-stakes domains like medicine and law. Metrics such as Expected Calibration Error (ECE) \citep{nixon2019measuring} and the Brier score \citep{glenn1950verification, guo2017calibration} quantify this misalignment; the Brier score in particular is a strictly proper scoring rule, so it rewards reported confidences that match empirical correctness and decomposes into calibration and sharpness terms, which is why it, rather than accuracy, is the natural target when the belief must drive the commit, verify, or abstain decision of \eqref{eq:bayes-decision}. Recent work explores reasoning-aware confidence estimators to improve calibration.

\noindent\textbf{Generalization to unseen domains.}
Generalization metrics evaluate whether reasoning ability transfers to domains or tasks that differ from the training data. Benchmarks such as BIG-Bench \citep{srivastava2023beyond}, MMLU \citep{hendrycks2020measuring}, and SuperGLUE \citep{wang2019superglue} provide diverse testbeds covering multiple knowledge areas, languages, and reasoning styles. Recent multilingual and multimodal evaluations \citep{bang2023multitask, zhang2023m3exam} extend this to cross-lingual and multimodal reasoning. These evaluations reveal systematic weaknesses, such as reasoning failures in low-resource languages or specialized professional tasks.

\noindent\textbf{Distribution-shift robustness.}
Another important dimension is robustness under distribution shifts, where test inputs come from distributions that differ from those of the training or prompting regimes. Recent studies \citep{liu2023good, yuan2023revisiting} show that even state-of-the-art LLMs experience sharp accuracy drops when evaluated on shifted domains, novel reasoning styles, or adversarial benchmarks. Distribution-shift evaluations expose brittleness in reasoning pipelines and highlight the need for continual adaptation or domain-robust training methods \citep{hendrycks2020pretrained}.

Human-in-the-loop evaluation complements automated metrics by letting experts inspect the process rather than only the outcome. Because the control-centric view treats intermediate state, validation, and recovery as first-class, human reviewers are well placed to catch failures that outcome accuracy hides, such as sound-looking but unfaithful rationales or errors the system never flagged through its own checks~\citep{natarajan2024humanintheloopaiintheloopautomatecollaborate}. In high-stakes domains such as medicine and education, keeping a human in the loop also yields a feedback signal that can be routed back into retrieval or training~\citep{MEMARIAN2024100053, shu2024human}. Such evaluation should report the proportion of errors identified and corrected, correction time, unnecessary intervention rate, and agreement among reviewers. Human--artificial intelligence (AI) system performance should also be compared with human-only and model-only baselines under matched time and cost, while monitoring reviewer workload and automation bias.

\section{Open Challenges and Future Directions}
\label{sec:challenges}
We close by revisiting the estimation-and-decision loop of Section~\ref{sec:framework} to ask, for each component, where control remains missing and where the field might go next. Read through the decompositions of \eqref{eq:bias-var-noise} and \eqref{eq:total-variance}, the open problems concentrate on the parts of the loop that current systems estimate least reliably: a state that is not yet a sufficient statistic, a transition mechanism whose bias is not repaired, an observation model whose likelihood is miscalibrated, a decision rule that cannot separate reducible from irreducible uncertainty, and effective dynamics that can amplify rather than attenuate errors under distribution shift. We group these into six research programs organized around the loop components rather than around individual methods: structured and faithful state; adaptive transition, search, and rollback; reliable validation; uncertainty-aware control and its compute budget; distributed and human-governed control; and the internalization of control together with its generalization under distribution shift.

\subsection{Structured State, Provenance, and Faithful Trace Interfaces}\label{sub:state-future}
The open problems for state representation are three statistical questions about the belief statistic. \emph{Sufficiency}: does the state carry the information the downstream decision will need, or is it lossy in the sense of Section~\ref{sub:state}? \emph{Compression}: how can irrelevant history be discarded without discarding decision-relevant information, so the statistic stays compact? \emph{Auditability}: is the externalized statistic faithful enough that validation and rollback can operate on it? The directions below, structured and typed state, belief-state abstraction, graph structure, predictive and latent state, and faithful trace interfaces, are answers to one or more of these three questions.

As discussed in Section~\ref{sub:state}, current methods for explicit state representation make intermediate reasoning observable by externalizing it into token-level traces, scratchpads, or branching structures. This is an important step, but the dominant interface remains a flat text buffer: states are only weakly typed, carry limited semantic annotation, rarely encode provenance or uncertainty explicitly, and do not track dependencies among intermediate conclusions.
A major future direction is therefore to move from \emph{state-as-text} toward \emph{state-as-object}: representations that are typed, queryable, updateable, and that support provenance tracking, uncertainty annotation, and dependency structure. Such state would allow downstream control components, including validation (Section~\ref{sub:validation}), constraint propagation, selective repair, and rollback (Section~\ref{sub:search}), to operate on well-defined objects rather than parsing unstructured text.

One immediate step beyond free-form traces is to externalize only the control-critical variables of reasoning, for example goals, constraints, retrieved evidence, commitments, uncertainty tags, and checkpoints, as typed slots or schema-backed objects rather than unstructured text. The aim is not to force every reasoning step into a rigid serialization, but to create a stable control surface for validation, provenance queries, and targeted state repair.
This selectivity matters because strict format requirements can degrade reasoning quality~\citep{tam2024let}, and recent analysis attributes this degradation primarily to the prompt level rather than the decoder, recommending that reasoning be decoupled from serialization~\citep{lee2026formattax}. Theoretical work confirms that overly restrictive output grammars reduce model expressivity, but also shows that alternating between unconstrained reasoning and constrained structured output can recover it~\citep{banerjee2025crane}.
This trade-off between representational richness and reasoning fidelity leads to a promising design principle that is to decouple reasoning from serialization: keep auxiliary deliberation flexible, but maintain a typed interface for the parts of state that later modules must inspect, validate, or revise.

\noindent\textbf{Belief-state abstraction and selective retention.}
Under partial observability, which is the default regime for LLM agents interacting with tools, users, and dynamic environments, conditioning on the full action--observation history often mixes task-relevant state with irrelevant detail, increasing both cost and brittleness~\citep{kaelbling1998planning}.
A more principled alternative is to maintain a compact \emph{belief state}: a sufficient statistic of prior interaction that separates stable task variables (e.g., goals, constraints, and progress markers) from transient observations and disposable deliberation. Recent work makes this idea concrete for LLM agents.
QuBE constructs a focused belief state via tool-enabled question answering to mitigate reasoning derailment under noisy, task-irrelevant observations~\citep{kim2024qube}.
PABU further models task progress explicitly and selectively retains only the essential interaction fragments needed for future decisions, substantially outperforming full-history conditioning across multiple agentic environments~\citep{jiang2026pabu}.
In multi-agent settings, CoBel-World extends belief modeling to collaborative reasoning by maintaining a shared belief world over both environment state and collaborators' mental states, reducing communication cost while improving coordination efficiency~\citep{wang2026collaborativebeliefreasoningllms}.

From a control perspective, compact belief states directly address trajectory bloat (Section~\ref{sec:uncertainty}) by discarding irrelevant history; the separation of stable, evidential, and ephemeral components clarifies which variables must be protected during rollback (Section~\ref{sub:search}) and which can be safely regenerated or discarded; and explicit belief uncertainty creates a natural substrate for uncertainty-aware transition selection (Section~\ref{sub:uncertainty}).

\noindent\textbf{Graph-structured relational state and dependency tracking.}
Neither linear traces nor branching trees represent the relational dependencies, entities, constraints, causal links, and intermediate conclusions, that arise in multi-hop reasoning. Graph-structured state makes these explicit and queryable, whether built from retrieved evidence~\citep{edge2024local, parekh2025structure} or maintained as agent memory updated online~\citep{anokhin2024arigraph, yang2026graph}. Treating the graph as the primary substrate of the state, rather than an external validation resource, serves \emph{auditability} and \emph{compression} directly: it supports localized rollback that invalidates only the affected subgraph (Section~\ref{sub:search}) and constraint propagation across related variables (Section~\ref{sub:validation}). Two problems remain: what structure to build is still largely hand-designed, with learned representation policies an early alternative~\citep{wu2025structure}, and current systems reach only workflow-level dependency tracking rather than classical truth maintenance, in which retracting one assumption invalidates all downstream conclusions~\citep{chang2025sagallm}.

\noindent\textbf{Multimodal state and world models.}
When CPS extends beyond text to robotics, embodied agents, and visual reasoning, the state must fuse modalities into one task-relevant representation, for example a serializable scene graph that focuses reasoning on task-relevant subsets~\citep{chen2026schema} or a belief updated from aggregated multimodal observations before acting~\citep{sun2024interactive}. World-model arguments make the stakes precise: flexible goal-directed behavior in effect requires a learned predictive model of the environment~\citep{richens2025general}, and coupling policy learning with learned world models enables lookahead~\citep{liu2026imagine,ha2018recurrent}. In the three questions, predictive state is a richer statistic that enables model-based search (Section~\ref{sub:adaptive-search}) and anticipatory validation, but its open problem is \emph{auditability}: calibrating and grounding it when imagined futures diverge from the real environment.

\noindent\textbf{Latent and continuous state representations.}
A complementary direction is to enrich the model's internal state rather than externalizing progressively more structure into text or graphs. Instead of expressing reasoning as discrete token-level traces, these methods operate in continuous latent space, increasing the computational bandwidth available per step while avoiding much of the token and serialization overhead of explicit chain-of-thought.
Concrete instances span pre-training and post-training: COCONUT feeds continuous hidden states back as input embeddings, enabling breadth-first-search-like exploration that outperforms discrete chain-of-thought on backtracking-heavy logical tasks~\citep{hao2025training}; Ouro scales the idea through a looped-transformer architecture whose modest-parameter LoopLMs match much larger standard transformers, an advantage attributed to superior knowledge manipulation rather than storage~\citep{zhu2025scaling}; and latent trajectories can be improved further by reinforcement learning with dense trajectory-level credit assignment~\citep{williams2026prioritize}. Theory supports this line, showing that looped transformers can simulate multi-step chain-of-thought through iterative latent-state refinement~\citep{saunshi2025reasoning}.

However, the control trade-off is clear. Latent state offers efficiency and expressive internal computation, but it also sacrifices the inspectability on which validation, rollback, and human oversight depend. For CPS settings, latent enrichment therefore appears most promising in hybrid architectures that externalize only control-critical variables, for example goals, constraints, commitments, and checkpoints, while allowing auxiliary computation to remain latent. This framing aligns directly with the central question of this survey: not whether reasoning is merely powerful, but whether it is controllable.

\noindent\textbf{The explicit--latent spectrum and cross-cutting open problems.}
Taken together, these directions suggest that state representation should be treated as a controllable design axis rather than a single fixed choice. At one end of the spectrum, fully explicit typed state maximizes observability, validation, rollback, and human oversight, but may impose reasoning and serialization costs. At the other end, highly latent state maximizes computational flexibility and efficiency, but weakens inspectability and external control. A central open problem is therefore to determine where along this spectrum a given task should operate, and how a system should move along the spectrum dynamically as its control requirements change. Stated as one design problem, the question is deciding what must be externalized for control, what can remain latent for efficiency, and how the externalized trace can remain faithful enough to support validation and rollback.

Several subordinate problems remain open. Current systems still provide only partial support for dependency-aware revision, selective invalidation, and principled rollback across structured reasoning state. Existing frameworks also remain fragmented: some emphasize belief compression, others graph structure, multimodal predictive state, or latent enrichment, but few combine these representational axes within a single coherent control architecture. Finally, the structure--reasoning trade-off remains unresolved at a general level. Two-pass architectures that decouple reasoning from serialization, selective externalization of control-critical variables, and learned representation policies are all promising, but comparative evidence across tasks and domains remains limited. Addressing these gaps will require treating state representation not as a passive container for reasoning, but as an active control surface of the reasoning system itself.

\noindent\textbf{Checkably faithful trace interfaces.}
A structured state interface is useful for oversight only if what it records is causally tied to the answer and can be verified step by step, which connects structured state directly to the problem of faithfulness.

A central challenge in LLM reasoning is ensuring that intermediate steps are faithful to the actual causal pathway leading to the final answer, rather than unfaithful rationalizations~\citep{turpin2023languagemodelsdontsay}. While reasoning enhancement techniques mentioned in Section~\ref{sec:taxonomy} often improve reasoning ability, they can also introduce unfaithful rationalizations~\citep{creswell2022faithfulreasoningusinglarge}. This problem undermines trust, as users may mistake fluent justifications for genuine reasoning, which is dangerous since unfaithful reasoning can hide systematic errors, propagate biases, and create a false sense of reliability in high-stakes applications~\citep{lightman2023let}.

We distinguish four recurring forms of unfaithfulness.

\textit{Answer-driven Rationalization:} LLM first settles on an answer and then fabricates a reasoning chain to justify it \citep{jin-etal-2025-disentangling-memory}. The answer can come from memory, prompt leakage, or superficial correlations~\citep{turpin2023languagemodelsdontsay}. For example, it produces the correct result $7 \times 8 = 56$ from memory, but presents a spurious derivation such as ``$7 \times 7 + 7 = 56$.''

\textit{Hallucinated Explanations:} LLM introduces an irrelevant theorem or non-existent concept into its reasoning chain, so the final answer may be correct while the supporting rationale is invalid, for example computing a right-triangle leg correctly but attributing the step to a fabricated ``Cosine Square Theorem'' rather than to the Pythagorean theorem \citep{lu2025auditingmetacognitivehallucinationsreasoning, cheng2025chainofthoughtpromptingobscureshallucination}.

\textit{Contradictory Reasoning:} LLM produces reasoning chains that contain internal inconsistencies, logical conflicts, or step-to-step contradictions that undermine the coherence of the explanation. For example, it first assumes ``if A then B'', later asserts ``A is true but B is false'', yet still concludes with the correct option \citep{liu2024selfcontradictoryreasoningevaluationdetection, gao2024dissectingdissonancebenchmarkinglarge}.

\textit{Bias-justifying Rationalization:} LLM's decision is influenced by hidden biases, but it provides a seemingly plausible explanation that masks this fact. For example, it favors the ``male engineer'' over the ``female engineer'' in a hiring scenario despite comparable backgrounds, claiming that the male description includes more skills \citep{wu2025doesreasoningintroducebias,echterhoff-etal-2024-cognitive}.

Several recent works have proposed methods to diagnose or mitigate these forms of unfaithfulness. For answer-driven rationalization, process supervision approaches such as the one studied by ~\citet{lightman2023let} explicitly supervise intermediate steps instead of only final answers, aiming to align the reasoning chain with the actual decision pathway. For hallucinated explanations, auditing frameworks~\citep{lu2025auditingmetacognitivehallucinationsreasoning} and controlled prompting strategies~\citep{cheng2025chainofthoughtpromptingobscureshallucination} have been developed to reduce reliance on fabricated concepts and to disentangle hallucinated rationales from correct outcomes. To address contradictory reasoning, specialized benchmarks and detection tools have been introduced. \citet{liu2024selfcontradictoryreasoningevaluationdetection} and \citet{gao2024dissectingdissonancebenchmarkinglarge} highlight logical incoherence and allow evaluation of consistency across steps. For bias-justifying rationalizations, recent studies~\citep{wu2025doesreasoningintroducebias, echterhoff-etal-2024-cognitive} investigate whether reasoning traces amplify social biases and propose debiasing techniques that enforce fairness-aware rationales.

Despite these efforts, none of the approaches fully eliminate unfaithfulness. Existing solutions often work only under restricted settings, fail to generalize across tasks, or introduce new trade-offs between faithfulness and performance. Ensuring that reasoning traces are consistently faithful remains an open challenge.

\noindent\textbf{From faithfulness to checkable interpretability.}
Faithfulness concerns whether an explanation is causally tied to the answer; interpretability concerns whether an observer can efficiently check each step. The two meet in designs that tie reasoning to verifiable structure. Process supervision trains with step-level signals rather than only final answers, and prover-verifier decompositions separate the generation of intermediate statements from an independent checker, yielding localized error signals and traces an auditor can follow~\citep{lightman2023let,uesato2022solving,kirchner2024prover}.

A complementary line grounds explanations in executable artifacts so that correctness is decided by running code or symbolic solvers rather than by judging prose. Program-Aided Language modeling and Program-of-Thoughts prompting express reasoning as short programs whose execution results become inputs to the next step; this both raises accuracy on math and data-processing tasks and makes the rationale inspectable through variables, control flow, and outputs \citep{gao2023pal,chen2023pot}. Beyond hard checks, process reward models score partial trajectories for logical soundness and task progress, enabling selection or reinforcement of higher quality reasoning without writing domain-specific verifiers for every step \citep{setlur2024rewarding}. Practical assessment combines step-level validity, entailment or contradiction tests within rationales, counterfactual perturbations that should flip the answer if a cited step is altered, and calibration metrics that compare stated confidence with actual correctness \citep{lanham2023measuring,lightman2023let}.

Open challenges remain. Faithful traces can be long and costly, while short traces can hide critical computations. Verifiers are fallible outside their domains, and process rewards risk rewarding verbosity over substance. Promising directions include structured rationales that commit to minimal sufficient evidence, hybrid traces that mix concise text with typed objects such as equations and tables, graph- or program-backed plans whose nodes are executable or checkable, and selective disclosure policies that provide full traces to auditors while summarizing salient steps for end users. The goal is a spectrum of transparency where every decisive operation is either executed, proven, or explicitly justified, and where models are trained and evaluated to keep their explanations causally tied to their answers \citep{kirchner2024prover,gao2023pal,setlur2024rewarding}.

\subsection{Adaptive Transition, Search, and Rollback Control}\label{sub:adaptive-search}
The open problems here concern making the controlled transition kernel $q(s_{t+1}\mid s_t,a_t)$, and the search over the belief it induces, adaptive rather than fixed. A hand-designed decomposition can contribute a bias term, and that bias can be repaired only if the controller detects it and changes how proposals are structured; symmetrically, the aggressiveness of search should respond to decision-relevant uncertainty rather than a preset schedule. What such adaptivity cannot remove is the risk it introduces when the feedback signal driving it is itself unreliable, which is the recurring theme below. The transition mechanisms of Section~\ref{sub:transition}, from decomposition and planning to program induction, largely rely on fixed, hand-designed structure: a prompt template, a decomposition schema, or a code-generation format chosen in advance and applied uniformly. This works when the structure matches the problem, but it transfers poorly when it does not, and it provides no way to revise the structure when intermediate evidence shows it to be wrong. A natural direction is to treat transition structure as itself controllable, adapting decomposition granularity, the degree of program grounding, and the planning horizon in response to the problem and to feedback from earlier steps. Early work points this way: adaptive decomposition expands structure only where intermediate states are judged complex~\citep{pandey2025adaptive}, and learned representation policies generate and adapt structural formats from reasoning demands rather than fixing them a priori~\citep{wu2025structure}. The open problem is a controller that decides how much structure to impose, and when to change it, rather than committing to a single transition style for an entire trajectory.

Adapting the structure of individual transitions extends naturally to adapting how alternatives are explored, compared, and revised.

From the control perspective developed in this survey, search in complex problem solving should
not be treated as a fixed procedure applied uniformly to all problems. Instead, search itself
constitutes a feedback-conditioned control process whose parameters, including branching factor, rollback depth,
expansion priority, and termination criteria, should be dynamically conditioned on feedback
signals from intermediate reasoning states.

Several recent methods instantiate fragments of this idea. Tree-of-Thoughts
\citep{yao2023tree} and its graph-based variants \citep{besta2024graph, bi2024forest}
introduce branching and rollback operators that enable recovery from early errors, providing
the structural component of search-path revision. However, their branching factors and
expansion rules remain fixed and do not adjust based on evolving uncertainty, so the
controller is not conditioned on intermediate feedback. Adaptive Graph of Thoughts
\citep{pandey2025adaptive} introduces dynamic decomposition and selectively expands reasoning
only when intermediate states are classified as complex, conditioning one control
parameter (decomposition granularity) on feedback, but leaving search dynamics such as
rollback scope and exploration breadth uncontrolled. Value model-guided search
\citep{yu2024ovm, feng2023alphazero} scores intermediate reasoning paths with a verifier and
retains only high-scoring candidates, implementing feedback-conditioned pruning. Yet recent
analyses reveal that such methods suffer from verifier failures at scale
\citep{yu2025scaling}, indicating that the feedback signal itself can be unreliable.
Uncertainty-Aware Value Models \citep{yu2025uncertainty} address this limitation by outputting
value distributions and using Group Thompson Sampling for branch selection, making the
controller sensitive to verifier reliability rather than treating verifier scores as ground
truth.

Despite these advances, no existing method fully closes the loop between search decisions and
the intermediate signals already available in the reasoning pipeline. Validation feedback
could narrow search when checks pass and trigger targeted rollback when violations are
detected; uncertainty estimates could widen exploration at high-uncertainty nodes and permit
commitment when confidence is sufficient. Realizing this vision requires reconceptualizing
search not as a static procedure layered on top of reasoning, but as a feedback-conditioned
controller whose structure, aggressiveness, and rollback behavior are continuously modulated
by the validation and uncertainty signals that the reasoning process already produces.

\subsection{Reliable Validators and Constraint Enforcement}
In the terms of this survey the validator supplies the action-conditioned observation model $p(o_{t+1}\mid s_{t+1},a_t)$, so its reliability is the question of whether that likelihood is well specified. A calibrated validator lets the update step correct bias and arrest state divergence, whereas a miscalibrated or gameable one is a misspecified likelihood that moves the belief in the wrong direction, and what no amount of validation can supply is a check on the parts of the task that lie outside the validator's domain. Validation is only as trustworthy as the validator. Executable and symbolic checkers give hard guarantees but cover a narrow slice of tasks, while learned validators such as process reward models and prover--verifier critics extend coverage to open-ended reasoning at the cost of reliability: they can be miscalibrated, exploited by the very policies they score, and brittle outside their training distribution. A central open direction is to make learned validators reliable enough to act as control gates, through better calibration, uncertainty over verifier scores, adversarial training against reward hacking, and validators that emit localized, checkable evidence rather than a single scalar~\citep{setlur2024rewarding, kirchner2024prover}. Recent analyses show that verifier reliability can degrade at scale~\citep{yu2025scaling}, motivating methods that model uncertainty over verifier scores instead of trusting them as ground truth~\citep{yu2025uncertainty}. Concretely, this points toward validators that report calibrated confidence rather than a point score, that localize which step or claim failed rather than judging a whole trace, that are trained adversarially against the policies they will gate, and that combine learned scores with hard executable or symbolic checks wherever those are available. Because validation, principled rollback, and calibration are, on our own account, the components that current systems control least, progress here is likely to matter more than further gains in proposal quality.

\subsection{Uncertainty-Aware Control and Compute Budgets}\label{sec:uncertainty}
This subsection treats uncertainty and compute as a single control problem, and three claims organize it. First, uncertainty estimates decide the control action: whether to answer, verify, branch, or abstain. Second, compute (tokens, samples, search depth, verifier calls) is the budget spent to carry out those decisions. Third, efficiency failures are what happen when compute is allocated without calibrated uncertainty, so effort is spent where confidence is already high and withheld where it is low.

The concrete open problem is estimation. To allocate compute by \eqref{eq:total-variance} the controller must estimate the epistemic component at each state, yet current models are overconfident and their token-level likelihoods align poorly with correctness \citep{liu2024uncertainty, zhang2025token}. Separating the reducible epistemic part from the irreducible aleatoric part is what tells the controller whether more compute will help or whether it should abstain \citep{huang2024survey}. Existing estimators, self-consistency, token-level entropy, Bayesian calibration, and Monte Carlo decoding, approximate a belief distribution over reasoning paths \citep{zhaoetal2025uncertainty}, but a per-state estimate of reducible uncertainty that is reliable enough to gate compute and abstention, and that stays calibrated under distribution shift, remains the central unmet need.

\noindent\textbf{Compute as a control budget.}
Uncertainty also governs how much computation a problem deserves, which turns efficiency into a control-budget question: effort should be spent where confidence is low and withheld where it is not.

Efficiency failures are, in this reading, the same thing as compute allocated without calibrated uncertainty. If the controller knew the epistemic component of \eqref{eq:total-variance} at each state, it would spend budget where that component is high and stop where it is low; without a calibrated estimate it cannot, and the two efficiency pathologies below are the two ways that misallocation shows up.

Reasoning overhead and search cost. When the controller cannot tell that uncertainty has already been resolved, it keeps spending: the trajectory degenerates into redundant, semantically overlapping steps that yield zero marginal reduction in the uncertainty of the output \citep{munkhbat2025selftrainingelicitsconcisereasoning,song2025prmbenchfinegrainedchallengingbenchmark,su2025tokenassortedmixinglatent, luo2025o1prunerlengthharmonizingfinetuningo1like, chen2025think23overthinkingo1like, li2025llmseasilylearnreason}. This is compute spent against a component that is no longer reducible, the efficiency analogue of the control mismatch in Table~\ref{tab:tab1}, and a calibrated stopping signal is exactly what would redirect it toward the decision points that still matter.

Trace switching and switching cost. The complementary failure is spending budget on premature moves between modes: lacking a calibrated estimate of which partial trajectory is more promising, the model switches between logical paths instead of committing \citep{wang2025thoughtsplaceunderthinkingo1like}, disrupting continuity, steering away from good solutions, and reducing accuracy \citep{yeo2025demystifyinglongchainofthoughtreasoning,yu2025dapoopensourcellmreinforcement,yuan2025whatspposcollapselongcot}.
In multi-agent frameworks, interference between competing traces further exacerbates this control failure, replacing structured traversal with scattered, high-cost jumps \citep{huang2024understandingplanningllmagents}.

\subsection{Distributed and Human-Governed Control}
Distributing control across agents and keeping a human in the loop both answer the same question of who governs which part of the trajectory. Read statistically, distributing control approximates the belief by an ensemble of local estimators, and a human in the loop adds an external, higher-quality observation and decision source; both change who supplies the estimate and the decision rather than adding a new component. The gain is variance reduction and broader coverage when the estimators are diverse, but what neither removes is a bias shared across correlated agents, the effective-sample-size collapse of \eqref{eq:ess} at the level of controllers rather than samples, nor a miscalibrated aggregation rule that rewards fluency over correctness. The underlying mechanisms of role specialization, communication protocols, and tool use were covered as a cross-cutting dimension in Section~\ref{sub:distributed}; the open problems here concern what remains uncontrolled when authority is shared rather than centralized.

Deliberation protocols further improve reliability when tasks are ambiguous or high-stakes. Debate and round-table consensus expose alternative rationales, reduce single-path bias, and enable aggregation rules that favor evidence and calibrated confidence rather than fluency \citep{irving2018ai,du2023improving,chen2023reconcile}. Long-horizon settings benefit from agent memories that record observations, reflections, and plans; Generative Agents demonstrate how retrieval from structured episodic memory supports coherent behavior over time, which in collaborative systems becomes an auditable log for human review \citep{park2023generative}. In practice, three safeguards help align collaboration with human goals: require evidence-linked messages with citations to artifacts produced by tools; attach confidence and uncertainty summaries to each proposal; and gate irreversible actions with reversible checkpoints that humans must approve.

Evaluation should reflect partnership rather than model-only accuracy. Useful measures include task success under time and cost budgets, correction counts and where they occurred in the workflow, stability across reformulations, agreement with expert judgment, and calibration of confidence against actual correctness. Known risks include persuasive but wrong rationales, echoing among similar agents, and human over-reliance. Mitigations pair dissenting critic roles with confidence-weighted voting, require verifiable intermediate artifacts, and favor diverse models or prompts in the team. The resulting blueprint treats the system as a cooperative ensemble where humans set objectives and constraints while agents decompose, propose, verify, and document reasoning in a way that remains inspectable and governable \citep{li2023camel,wu2024autogen,chen2023reconcile}.

\subsection{Internalized Control and Generalization under Shift}
A complementary direction to external scaffolding is to internalize control into the model's own parameters, so that behaviors such as validating a step, maintaining a plan, or abstaining under uncertainty are produced by the policy rather than imposed from outside. As Section~\ref{sub:training} put it, this amortizes the loop's components, the proposal kernel, the observation model, and the decision rule, into the weights, so the open question is not whether control can be internalized but whether these amortized estimators stay well specified once the input distribution shifts. An amortized likelihood that was calibrated in training can become misspecified off-distribution, at which point the effective error dynamics in \eqref{eq:error-recursion} may become less stable or locally expansive and internalized control can degrade silently, a failure that external validation may help expose. Supervised fine-tuning on process-annotated traces, preference optimization, and reinforcement learning with process-level rewards each move some control inward, and reasoning-oriented training has been shown to elicit more explicit multi-step behavior when that behavior is rewarded~\citep{guo2025deepseek, setlur2024rewarding}. The open questions concern reliability rather than feasibility. Internalized control can be unstable under distribution shift, hard to audit because it leaves no external trace, and vulnerable to reward hacking when the training signal is imperfect. The central challenge is to internalize control in a way that remains verifiable and that degrades gracefully, rather than silently, once the model leaves its training distribution.

Whether control internalized during training survives outside the training distribution is itself the question of generalization.

Generalization to novel problem types tests whether LLMs can remain effective when the task structure, domain assumptions, or required reasoning style differ from those seen during pretraining or adaptation \citep{srivastava2023beyond, guoDomainAdaptationGeneralization2022}. Prior work has examined this issue through zero-shot transfer, in-context adaptation, compositional generalization, and cross-domain transfer \citep{wei2022finetuned, kojima2022large, dongSurveyIncontextLearning2024, zhou2023leasttomostpromptingenablescomplex, chen2023pot, sanh2022multitask_prompted_training, JMLR:v25:23-0870}. From the CPS perspective, these settings can be viewed as instances of structural shift: the key question is whether the control components introduced in Section~\ref{sec:taxonomy} still regulate reasoning when familiar problem formats no longer apply.

Generalization failures often arise because the model's control structure does not transfer with the surface task performance. When the problem structure changes, the model may construct inappropriate intermediate states, apply familiar transition patterns to incompatible reasoning demands, or continue along invalid trajectories because validation and uncertainty signals no longer diagnose errors reliably. Weak search, rollback, and iterative correction mechanisms further amplify these failures, since early mistakes cannot be easily revised once the reasoning trajectory has moved in the wrong direction.
Thus, strong performance on seen benchmarks does not necessarily imply robust reasoning on novel problem types. The core CPS challenge is not only whether task knowledge transfers, but whether control over state construction, transition, validation, and recovery remains effective under structural shift. Robust CPS therefore requires control mechanisms that generalize beyond familiar problem formats, rather than reasoning traces that merely appear plausible within known distributions.

\section{Conclusion}
This survey has argued that progress on complex problem solving with large language models is, to a greater degree than is usually recognized, a statistical question of estimation and decision rather than of raw reasoning ability alone. We read CPS through an action-conditioned state-space scaffold: reasoning supplies candidate moves, control shapes and evaluates those proposals, selected actions alter subsequent state transitions or observations, evidence updates a belief over latent solution state, and a decision rule determines whether to commit, verify, branch, roll back, or abstain. This view organizes a wide range of methods into a small set of components: externalized control-relevant state as an interface for estimation and intervention; transition structuring as proposal shaping; validation as an observation or evidence model; search and rollback as mechanisms for preserving and revisiting competing trajectories; and uncertainty management as decision-relevant belief information used together with the loss. Distributed and training-time control remain cross-cutting implementation choices. Three themes recur. First, many techniques presented as reasoning improvements are, on closer inspection, mechanisms that shape proposals, estimate state or evidence, search over alternatives, or implement decision rules. Second, validation, principled rollback, and calibrated uncertainty are among the weakest current components; they can reduce injected error, expose unstable trajectories, and support recovery, but none by itself guarantees that the effective dynamics in \eqref{eq:error-recursion} are contractive. Third, problem--control fit is best read as a diagnostic hypothesis: observed failure signatures suggest, but do not uniquely identify, the underlying error source, and an intervention is useful only to the extent that it acts on the source actually responsible. In particular, repeated sampling can reduce Monte Carlo or sampling variance without correcting a shared systematic bias or irreducible ambiguity. The practical implication is therefore diagnostic rather than prescriptive: before increasing reasoning effort, determine which control bottleneck is most plausibly implicated and test whether the corresponding intervention repairs it under a comparable budget.

These observations suggest a small set of design principles for statistical control of CPS systems. First, externalize only the control-critical parts of state, such as goals, constraints, commitments, and checkpoints, rather than assuming that every reasoning token is either necessary or faithful. Second, validate transitions before committing to them when a reliable check is available, so that errors can be caught close to the step that produces them. Third, use calibrated, decision-relevant uncertainty to allocate search and verification budget rather than treating low confidence as a complete decision rule by itself. Fourth, make rollback targeted, invalidating only the affected part of the trajectory rather than restarting from scratch. Fifth, evaluate the process, not only final accuracy, and use matched-budget interventions when testing the problem--control-fit hypothesis. Taken together, these principles restate the survey's central claim as concrete guidance: treat state, proposal shaping, validation, search, and uncertainty as components of a designable estimation-and-decision system, rather than treating longer generation as a substitute for statistical control.

\clearpage
\appendix
\section{Method--Mechanism Detail}\label{app:mechanisms}
This appendix refines the checkmark mapping of Table~\ref{tab:tab2} for the most representative methods, giving the mechanism each uses and what it leaves uncontrolled.

\begin{table}[h!]
\centering
\footnotesize
\setlength{\tabcolsep}{4pt}
\caption{A closer view of representative methods: the component each implements, the mechanism it uses, and what it leaves uncontrolled. This refines the checkmark mapping of Table~\ref{tab:tab2} into the argument the survey makes about each pattern; evaluation signals for each component are collected separately in Table~\ref{tab:tab5}.}
\label{tab:tab6}
\begin{tabularx}{\textwidth}{@{}p{1.9cm}p{2.0cm}XX@{}}
\toprule
Method & Component & Mechanism & What remains uncontrolled \\
\midrule
Chain-of-Thought & Explicit state & externalizes a linear textual trace & validity of the individual transitions \\
Tree-of-Thoughts & State, search & branches over candidate states with backtracking & branching and rollback are fixed, not feedback-driven \\
PAL / PoT & Transition, validation & offloads computation to executable code & correctness of the problem formulation \\
Self-Refine / Reflexion & Search, rollback & revises a draft using self-generated feedback & reliability of the self-critique \\
Verifier-guided search & Validation, uncertainty & scores partial paths with a verifier and prunes & verifier failures at scale \\
Multi-agent & Distributed control & splits roles across agents with cross-checks & correlated errors across agents \\
\bottomrule
\end{tabularx}
\end{table}
\clearpage

\section*{Statements and Declarations}

\subsection*{Funding}

This work was partially supported by the U.S. National Science Foundation (NSF)
[DMS-2124493, DMS-2311297, DMS-2319279, DMS-2318809]
and the National Institutes of Health (NIH) [R01GM152814].

\subsection*{Competing Interests}

The authors declare that they have no competing interests.

\bibliography{ref}

\end{document}